\documentclass[lettersize,journal]{IEEEtran}

\usepackage{afterpage}
\usepackage{amssymb}
\usepackage{amsmath,amsfonts}
\usepackage{physics}
\DeclareMathOperator*{\minimize}{minimize}
\usepackage{bbm}
\usepackage{dsfont}
\usepackage{algorithm}
\usepackage{algorithmic}
\usepackage{subfigure}
\usepackage{subcaption}
\usepackage{float}
\usepackage{setspace}
\usepackage{stmaryrd}
\usepackage{graphicx}
\usepackage{mwe}
\usepackage{array}
\usepackage{multirow}
\usepackage{multicol}
\usepackage[font=footnotesize]{caption}
\usepackage{textcomp}
\usepackage{stfloats}
\usepackage{url}
\usepackage{lettrine}
\usepackage{verbatim}
\usepackage{graphicx}
\usepackage{cite}
\usepackage{ifpdf}
\usepackage{array}
\usepackage[skins]{tcolorbox}
\usepackage{tabularx,array,booktabs}
\usepackage{setspace}
\usepackage{xcolor}
\usepackage{bm}
\usepackage{amsthm}
\usepackage{mathtools, nccmath}
\usepackage{chngcntr}
\usepackage{comment}
\usepackage{lipsum}

\DeclareMathOperator*{\argmin}{arg\,min}

\usepackage{textcase}

\theoremstyle{plain}

\newenvironment{proofofsketch}{%
\proof}{\endproof}
\def\BibTeX{{\rm B\kern-.05em{\sc i\kern-.025em b}\kern-.08em
    T\kern-.1667em\lower.7ex\hbox{E}\kern-.125emX}}
\allowdisplaybreaks
\graphicspath{{Figures/}}
\begin{document}
\title{ASFL: An Adaptive Model Splitting and Resource Allocation  Framework for Split Federated Learning}
\author{\IEEEauthorblockN{Chuiyang Meng\IEEEauthorrefmark{1}, \textit{Graduate Student Member, IEEE}, Ming Tang\IEEEauthorrefmark{2}, \textit{Member, IEEE}, and Vincent W.S. Wong\IEEEauthorrefmark{1}, \textit{Fellow, IEEE}}\\
\thanks{\IEEEauthorblockA{\IEEEauthorrefmark{1}Department of Electrical and Computer Engineering, The University of British Columbia, Vancouver, Canada}\\
\IEEEauthorblockA{\IEEEauthorrefmark{2}Department of Computer Science and Engineering, Southern University of Science and Technology, China}\\
E-mail: {\{chuiyangmeng, vincentw\}@ece.ubc.ca}, tangm3@sustech.edu.cn}}
\maketitle

\begin{abstract}
Federated learning (FL) enables multiple clients to collaboratively train a machine learning model without sharing their raw data. 
However, the limited computation resources of the clients may result in a high delay and energy consumption on training. 
In this paper, we propose an adaptive split federated learning (ASFL) framework over wireless networks.
ASFL exploits the computation resources of the central server to train part of the model and enables adaptive model splitting as well as resource allocation during training.
To optimize the learning performance (i.e., convergence rate) and efficiency (i.e., delay and energy consumption) of ASFL, we theoretically analyze the convergence rate and formulate a joint learning performance and resource allocation optimization problem. 
Solving this problem is challenging due to the long-term delay and energy consumption constraints as well as the coupling of the model splitting and resource allocation decisions.
We propose an online optimization enhanced block coordinate descent (OOE-BCD) algorithm to solve the problem iteratively.
Experimental results show that when compared with five baseline schemes, our proposed ASFL framework converges faster and reduces the total delay and energy consumption by up to 75\% and 80\%, respectively.
\end{abstract}


\section{Introduction}
\label{sec:introduction}


{The emergence of federated learning (FL) \cite{mcmahan2017communication} enables distributed model training across clients (e.g.,  smartphones, sensors, and embedded systems) while preserving data privacy. 
However, in practical systems, conventional FL suffers from several limitations. 
First, FL requires clients to train the entire model locally, which imposes substantial computation and energy burden on resource-constrained clients. 
As the model and datasets grow in size and complexity, training full models on these clients becomes prohibitively expensive in terms of the delay and energy consumption on training.
This degrades the learning efficiency significantly. 
Second, the central server remains under-utilized in FL, as it only performs simple parameter aggregation with a small computation load. 
This can lead to system-wide inefficiency, especially in scenarios where the central server is equipped with powerful computation resources (e.g., graphics processing units (GPUs)) and an uninterrupted power supply.}

Recently, split learning (SL) \cite{gupta2018distributed} has been proposed to address the aforementioned issues.
During SL training, the model is split into two parts.
Each client maintains one part of the model (called the client-side model), while the central server stores the remaining part (called the server-side model).
The client performs forward propagation (FP) and determines the intermediate output using its client-side model and local data. 
The intermediate output is sent to the central server.
The central server continues the training with backpropagation (BP) using the server-side model and determines the gradient of the model.
It sends the gradient back to the client to complete the BP. 
When a client has completed its BP with its client-side model, it then transfers the model to the next client for further training.
{In \cite{vepakomma2018split}, the authors applied SL to the use case of eHealth.
In \cite{poirot2019split}, the authors utilized SL for medical image classification.
In \cite{li2023convergence}, the authors provided theoretical analysis and derived the convergence bound of sequential SL.
In \cite{zhang2025split}, the authors proposed a hierarchical SL scheme for fine‑tuning large language models over wireless networks.  
In \cite{lin2025leo}, the authors adopt SL for satellite networks with intermittent connectivity and resource constraints.} 
Although SL can leverage the computation resources of the central server and has attracted significant attention in the research community
\cite{vepakomma2018split, poirot2019split, li2023convergence, zhang2025split, lin2025leo},
it has two limitations.
First, this sequential learning approach can cause the model to forget previously learned information when new information is available, which can lead to catastrophic forgetting \cite{kirkpatrick2017overcoming}.
Second, a client may experience a long idle time while waiting for other clients to finish their training.
This can reduce the learning efficiency.

To address the aforementioned limitations, split federated learning (SFL) has been proposed \cite{thapa2022splitfed, hong2022efficient, 10234718, 10378869, han2024convergence, dachille2024impact}. 
In SFL, in addition to model splitting as in SL, clients can train their client-side models in parallel.
In \cite{thapa2022splitfed}, the authors proposed an integrated learning framework of SL and FL.
In \cite{hong2022efficient}, the authors proposed a split-mix strategy for FL to achieve model customization.
In \cite{10234718}, the authors proposed an SFL framework for a ring topology.
In \cite{10378869}, the authors proposed a knowledge distillation approach tailored for personalized federated split learning.
{In \cite{han2024convergence}, the authors analyzed the convergence rate of SFL under strongly convex, convex, and non-convex settings. 
In \cite{dachille2024impact}, the authors theoretically and empirically analyzed the effect of model splitting points on the learning performance of SFL.}
However, the aforementioned works \cite{thapa2022splitfed, hong2022efficient, 10234718, 10378869, han2024convergence, dachille2024impact} did not consider the physical layer characteristics (e.g., wireless channel conditions) when implementing SFL in practical wireless systems.
{Unlike the conventional FL approaches where model exchange occurs after several model updates, SFL requires transmitting the intermediate output and gradient in each model update.
It is crucial to allocate network resources efficiently in order to improve the learning efficiency. }

Some recent works have proposed resource allocation algorithms for SFL \cite{10274134, 10314792, 10301639, 10040976, 10304624, lin2023efficient, tirana2024workflow, lin2024adaptsfl, 10740645, 10910050, 10700751, 10855336, lin2025hasfl, 10980018}.
In \cite{10274134}, the authors proposed a clustering-based SFL framework for model splitting and resource allocation.
In \cite{10314792}, the authors proposed a personalized SFL framework in wireless networks.
The authors in \cite{10301639} proposed a scalable SFL framework which balances the learning performance and delay on training.
The authors in \cite{10040976} proposed a cluster-based parallel SL framework and a resource allocation strategy.
In \cite{10304624}, the authors proposed an SFL framework to jointly determine the cut layer and bandwidth allocation for clients.
We denote this framework as ACC-SFL.
In \cite{lin2023efficient}, the authors proposed an efficient parallel SL (EPSL) framework which minimizes the training latency.
The authors in \cite{tirana2024workflow} explored the integration of multiple helpers in parallel SL and proposed a workflow scheduling algorithm. 
The authors in \cite{lin2024adaptsfl} proposed an SFL framework which adaptively controls the model splitting and client-side model aggregation in SFL.
The authors in \cite{10740645} proposed a fine-grained parallelization framework to accelerate SFL on heterogeneous clients.
The authors in \cite{10910050} proposed a wireless SFL framework which jointly tackles the data heterogeneity and client heterogeneity.
{The authors in \cite{10700751} optimized the model splitting points to minimize the overall training latency in SFL.
The authors in \cite{10855336} proposed a split federated low-rank adaptation framework to fine-tune large language models in wireless networks.
The authors in \cite{lin2025hasfl} adaptively controlled the batch sizes and model splitting points for edge devices in wireless networks to tackle the resource heterogeneity issue.
The authors in \cite{10980018} proposed a two-tier hierarchical SFL framework in wireless networks and optimized its resource allocation.}

The aforementioned works \cite{10274134, 10314792, 10301639, 10040976, 10304624, lin2023efficient, tirana2024workflow, lin2024adaptsfl, 10740645, 10910050, 10700751, 10855336, 10980018, lin2025hasfl} assumed that there are no errors in the received data.
In practical wireless systems, channel fading can lead to packet errors during communications, which can degrade the learning performance.
Moreover, these works assumed that the models are pre-split before training and require periodic client-side model aggregation by the central server, resulting in significant communication overhead.
They do not consider adaptive model splitting, where the models can be split at different layers in each training round and only several middle layers are transmitted between the clients and the central server, which can potentially reduce the communication overhead.

In this paper, we address the following question:
\textit{How to jointly improve the learning performance and reduce the delay and energy consumption of SFL training over wireless networks?}
We aim to develop an SFL framework which enables adaptive model splitting and resource allocation over wireless networks.
Achieving this goal is challenging due to the following reasons.
First, the dynamic wireless channel conditions may introduce packet errors and thus degrade the learning performance.
Second, it is challenging to characterize the effect of model splitting and resource allocation decisions (i.e., resource block (RB) and transmit power allocation decisions) on the convergence rate.
Third, the coupling of those decisions further complicates the solution. 
To overcome these challenges, our work makes the following contributions:

\begin{itemize}
    \item {We propose an adaptive SFL (ASFL) framework to determine the model splitting and resource allocation decisions in each training round.
    The proposed ASFL framework can improve the convergence rate and adapt to dynamic channel conditions in wireless networks.
    In addition, ASFL takes into account packet errors during the intermediate output transmission and their impact on the learning performance.}

    \item We analyze the convergence rate of ASFL.
    In particular, we quantify the impact of those decisions (i.e., model splitting, RB allocation, and transmit power allocation decisions) on the convergence rate.
    Based on the analytical results, we formulate a joint problem which optimizes the learning performance subject to the long-term delay and energy consumption constraints of ASFL.

    \item To solve the formulated problem, we propose an online optimization enhanced block coordinate descent (OOE-BCD) algorithm by decoupling the problem into a model splitting subproblem, an RB allocation subproblem, and a transmit power allocation subproblem.
    To solve the model splitting subproblem, we propose an online optimization algorithm to determine the model splitting decision by considering the long-term delay and energy consumption constraints.
    The stability of our proposed online optimization algorithm is guaranteed by showing that the decisions satisfy the long-term delay and energy consumption constraints.
    Then, we determine the RB allocation decision by solving an integer programming problem.
    Finally, we propose an iterative algorithm to solve the transmit power allocation subproblem.
    We solve these subproblems alternately in each training round.

    \item We conduct experiments on CIFAR-10 and CIFAR-100 datasets using VGG-19 and ResNet-50.
    We compare our proposed ASFL framework with federated averaging (FedAvg) \cite{mcmahan2017communication}, SL \cite{vepakomma2018split}, SFL \cite{thapa2022splitfed}, ACC-SFL \cite{10304624}, and EPSL \cite{lin2023efficient}. 
    Results show that in a wireless network setting with 10 clients, our proposed ASFL framework converges faster than the baseline schemes and can reduce the total delay and energy consumption on training by up to 75\% and 80\%, respectively. 
\end{itemize}

The rest of this paper is organized as follows.
The system model is introduced in Section \ref{sec:system_model}.
The theoretical analysis and problem formulation are presented in Section \ref{sec:theoretical_analysis}.
In Section \ref{sec:methodology}, we present our proposed OOE-BCD algorithm.
Experimental results are given in Section \ref{sec:performance_evaluation}.
Conclusions are drawn in Section \ref{sec:conclusion}.

\textit{Notations}: We use italic upper case letters, boldface upper case letters, and boldface lower case letters to denote sets, matrices, and vectors, respectively. 
$\mathbb{R}^{M \times N}$ denotes the set of $M \times N$ real-valued matrices.
$\mathbf{1}_{N} \in \mathbb{R}^{N}$ and $\mathbf{0}_{N} \in \mathbb{R}^{N}$ denote the all-ones and all-zeros column vectors with dimension $N$, respectively.
Mathematical operators 
$\mathbb{E}[\cdot]$, $(\cdot)^{\intercal}$, $[\cdot;\cdot]$, $\langle\cdot\rangle$, $\lvert\cdot\rvert$, $\Vert\cdot\Vert_{0}$, and $\Vert\cdot\Vert$ denote the expectation, transpose, column-wise concatenation, inner product, absolute value, 0-norm, and 2-norm, respectively.
$\sim$ denotes ``distributed as''.
$\mathcal{U}[\cdot]$ and $\mathcal{CN}(\cdot)$ denote the uniform distribution and complex normal distribution, respectively.
The key notations used in this work are summarized in Table \ref{tb:central_tb}.

\begin{table*}[t]
\centering
\caption{List of key notations}
\label{tb:central_tb}

\begingroup
\small
\setlength{\tabcolsep}{8pt}        
\renewcommand{\arraystretch}{1.08} 
\begin{spacing}{0.95}              

\begin{tabular}{|>{\centering\arraybackslash}m{0.12\linewidth}|
                    >{\centering\arraybackslash}m{0.3\linewidth}|
                    >{\centering\arraybackslash}m{0.12\linewidth}|
                    >{\centering\arraybackslash}m{0.3\linewidth}|}
\hline
\textbf{Notation} & \textbf{Definition} & \textbf{Notation} & \textbf{Definition} \\
\hline
$B$ & Bandwidth of a resource block & $q_{m}$ & Output size of the $m$-th layer (in bits) \\
\hline
$B^{\mathrm{DN}}$ & Downlink bandwidth & $\mathcal{R}$ & Set of all training rounds \\
\hline
$c_{n}^{\mathrm{DN}, r}$ & Downlink transmission rate for client $n$ in the $r$-th training round & $R$ & Number of training rounds \\
\hline
$c_{n}^{\mathrm{UP}, r}$ & Uplink transmission rate for client $n$ in the $r$-th training round & $s_{n}^{r}$ & Packet error rate of the intermediate output transmission of client $n$ in the $r$-th training round \\
\hline
$D_{n}$ & Number of training samples of client $n$ & $u_{n}^{r}$ & Allocation decision of the $k$-th RB for client $n$ in the $r$-th training round \\
\hline
$f_{n}$ & CPU frequency of client $n$ & $v_{m}^{\mathrm{FP}}$ & Computation workload for FP of the $m$-th layer (in FLOPs) \\
\hline
$f^{\mathrm{s}}$ & CPU frequency of the central server & $v_{m}^{\mathrm{BP}}$ & Computation workload for BP of the $m$-th layer (in FLOPs) \\
\hline
$\left\lvert h_{n}^{r}\right\rvert$ & Channel gain for client $n$ in the $r$-th training round & $\mathbf{w}_{n}^{r}$ & Client $n$'s model in the $r$-th training round \\
\hline
$\mathcal{M}$ & Set of all model layers & $\mathbf{z}_{n}^{r}$ & Intermediate output of client $n$ in the $r$-th training round \\
\hline
$\mathcal{K}$ & Set of all RBs & $\alpha$ & Waterfall threshold \\
\hline
$K$ & Number of RBs & $\beta_{n}^{r}$ & Binary indicator of packet error experienced by client $n$ in the $r$-th training round \\
\hline
$M$ & Number of model layers & $\gamma$ & Upper bound of average delay on training \\
\hline
$\mathcal{N}$ & Set of all clients & $\delta$ & Upper bound of average energy consumption on training \\
\hline
$N$ & Number of clients & $\lambda_{m}^{r}$ & Model splitting decision of the $m$-th layer in the $r$-th training round \\
\hline
$N_{0}$ & Received noise power spectral density & $\bm\xi_{n}$ & A mini-batch of training samples of client $n$ \\
\hline
$p_{n}^{r}$ & Transmit power for client $n$ in the $r$-th training round & $\psi_{m}$ & Size of the $m$-th layer (in bits) \\
\hline
$p^{\mathrm{s}}$ & Transmit power for the central server & $\kappa^{\mathrm{c}}$ & Number of CPU cycles required for a client to complete one FLOP \\
\hline
$p^{\mathrm{max}}$ & Maximum transmit power & $\phi$ & Energy consumption coefficient \\
\hline
\end{tabular}

\end{spacing}
\endgroup
\end{table*}

\section{System Model}
\label{sec:system_model}
We consider SFL over a wireless network.
Let $\mathcal{R}=\{1,2,\ldots, R\}$ and $\mathcal{N} = \{1, 2, \ldots, N\}$ denote the set of $R$ training rounds and the set of $N$ clients, respectively.
Each client $n \in \mathcal{N}$ has a local dataset $\mathcal{D}_{n}$ with $D_{n}$ training samples.
Let $\mathbf{w}_{n}^{r}$ denote client $n$'s model in the $r$-th training round.
We consider the model has $M$ layers.
The set of layers is denoted by $\mathcal{M} = \{1, \ldots, M\}$.
In the $r$-th training round, $\mathbf{w}_{n}^{r}$ is split into a client-side model $\mathbf{w}_{n}^{\mathrm{c}, r}$ and a server-side model $\mathbf{w}_{n}^{\mathrm{s}, r}$, which are trained by client $n$ and the central server, respectively.
We have $\mathbf{w}_{n}^{r} = [\mathbf{w}_{n}^{\mathrm{c}, r}; \mathbf{w}_{n}^{\mathrm{s}, r}]$.

For client $n \in \mathcal{N}$, let $\bm{\xi}_{n} = (\mathbf{x}_{n}, \mathbf{y}_{n}) \sim \mathcal{D}_{n}$ denote a mini-batch of training samples of $\mathcal{D}_{n}$, where $\mathbf{x}_{n}$ and $\mathbf{y}_{n}$ are the training data and the corresponding labels, respectively.
Let $F_{n}(\mathbf{w}_{n}^{r}; \bm{\xi}_{n})$ denote the local loss function of client $n$ on  $\bm{\xi}_{n}$ with model $\mathbf{w}_{n}^{r}$.
We denote the expected loss of client $n$ with model $\mathbf{w}_{n}^{r}$ as $F_{n}(\mathbf{w}_{n}^{r}) = \mathbb{E}_{\bm{\xi}_{n} \sim \mathcal{D}_{n}}F_{n}(\mathbf{w}_{n}^{r}; \bm{\xi}_{n})$.
Let $\bar{\mathbf{w}}^{r}$ denote the average model of all clients at the beginning of the $r$-th training round, which is given by
$\bar{\mathbf{w}}^{r} = \sum_{n=1}^{N} D_{n} \mathbf{w}_{n}^{r}/\sum_{j=1}^{N} D_{j}$.
The corresponding global loss function is denoted as  $F(\bar{\mathbf{w}}^{r}) = \sum_{n\in\mathcal{N}} D_{n}F_{n}(\mathbf{w}_{n}^{r})/\sum_{j\in\mathcal{N}} D_{j}.$
We aim to minimize $F(\bar{\mathbf{w}}^{R+1})$ after $R$ training rounds.
The optimal model is denoted as $\mathbf{w}^{\star}$.

\subsection{Learning Model}
\label{sec:learning_model}


\begin{figure*}[t]
    \centering
    \includegraphics[width=0.8\textwidth]{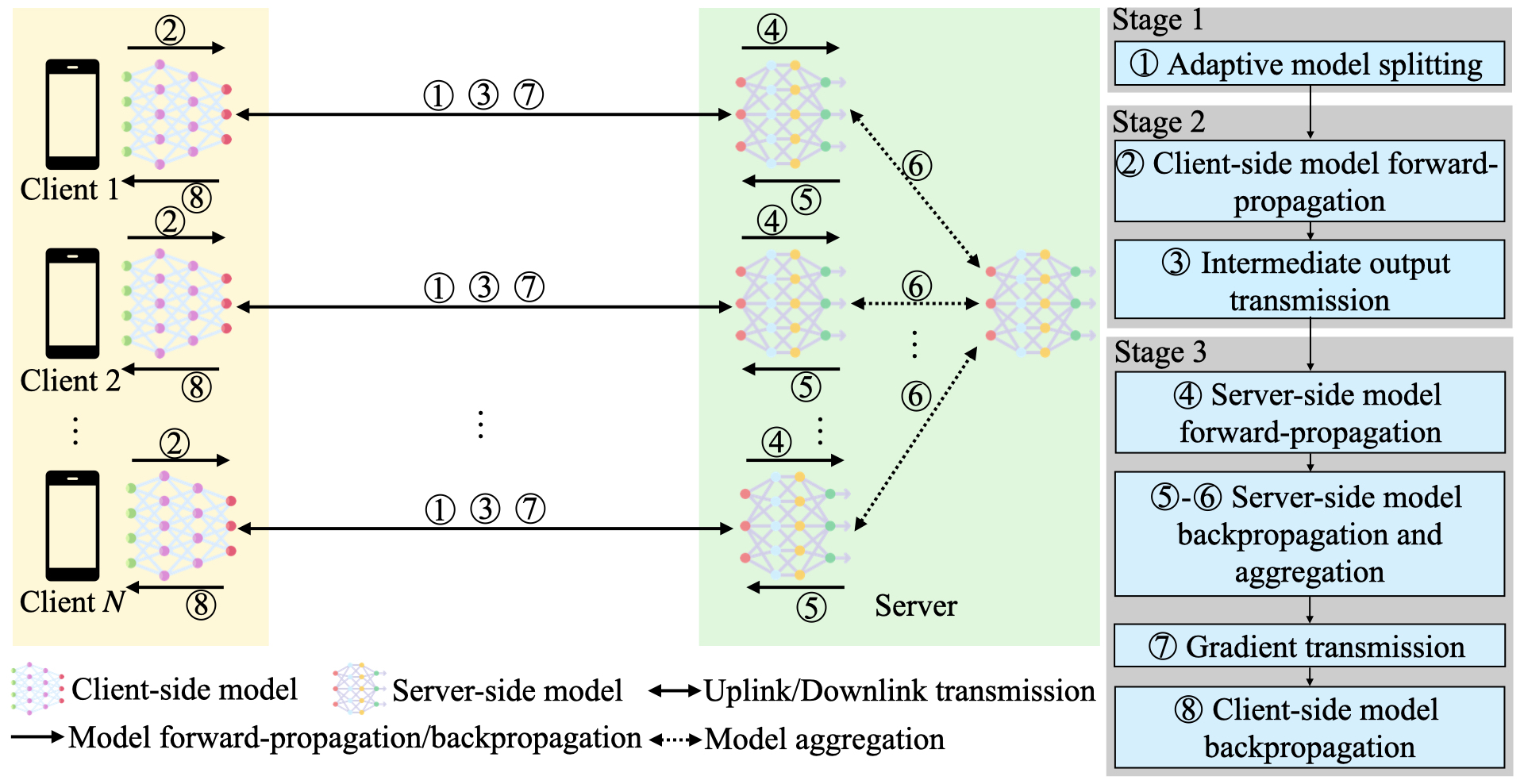}
    \caption{Illustration of our proposed ASFL framework.}
    \label{fig:system_model}
\end{figure*}
\begin{figure}[t]
\centering
    \includegraphics[width=0.48\textwidth]{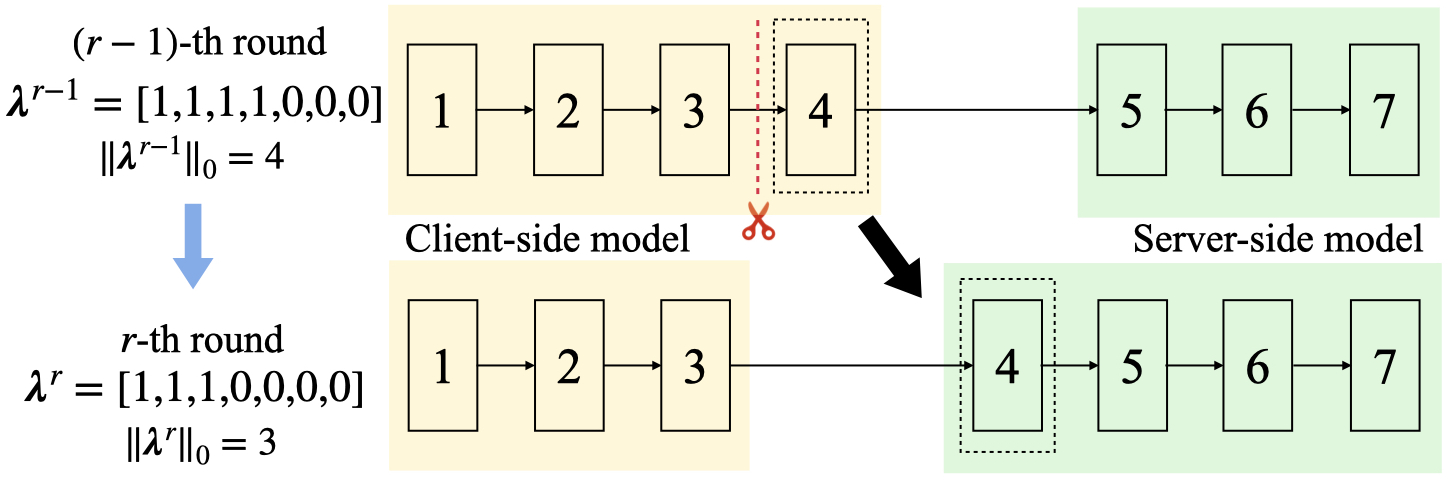}
    \caption{An example of the adaptive model splitting process in stage 1 for a model with 7 layers. 
    A client sends its fourth layer to the central server at the beginning of the $r$-th training round.}
    \label{fig:MS}
\end{figure}
An illustration of our proposed ASFL framework is shown in Fig. \ref{fig:system_model}.
The proposed framework has three stages.

\textbf{Stage 1}: 
An adaptive model splitting process is invoked.
Let $\boldsymbol{\lambda}^{r} = (\lambda_{1}^{r}, \lambda_{2}^{r}, \ldots, \lambda_{M}^{r}) \in \{0,1\}^{M}$ denote the model splitting decision vector in the $r$-th training round.
If $\lambda_{m}^{r} = 1$, then the $m$-th layer is for the client-side model. 
Otherwise, it is for the server-side model.
Each client has at least one layer in its client-side model for privacy protection, i.e., $\lambda_{1}^{r} = 1$. 
Note that $\Vert\boldsymbol{\lambda}^{r}\Vert_{0}$ denotes the number of layers of the client-side model.
If $\Vert\boldsymbol{\lambda}^{r}\Vert_{0} < \Vert\boldsymbol{\lambda}^{r-1}\Vert_{0}$, then the clients have fewer layers in the $r$-th training round than in the $(r-1)$-th training round.
Hence, the clients transmit the model parameters of middle layers (i.e., from the $(\Vert\boldsymbol{\lambda}^{r}\Vert_{0}+1)$-th to the $\Vert\boldsymbol{\lambda}^{r-1}\Vert_{0}$-th layer) to the central server.
Otherwise, the central server transmits the model parameters of middle layers to the clients. 
In addition, we require that all layers up to and including the $\Vert\boldsymbol{\lambda}^{r}\Vert_{0}$-th layer are for the client-side model, while the subsequent layers are for the server-side model.
Therefore, we have $\lambda_{m}^{r} \leq \lambda_{m-1}^{r}$, for $m \in \mathcal{M}\backslash\{1\}$.
An example of adaptive model splitting is shown in Fig. \ref{fig:MS}.
Based on the model splitting decision $\boldsymbol{\lambda}^{r}$, we denote the client-side model and server-side model of client $n \in \mathcal{N}$ at the beginning of the $r$-th training round as $\mathbf{w}_{n, \boldsymbol{\lambda}^{r}}^{\mathrm{c}, r}$ and $\mathbf{w}_{n, \boldsymbol{\lambda}^{r}}^{\mathrm{s}, r}$, respectively.

\textbf{Stage 2}: 
Each client performs FP using its client-side model and local dataset.
For client $n \in \mathcal{N}$, it generates the intermediate output, which is denoted as $\mathbf{z}_{n}^{r}$.
Let $\Phi_{n}^{\mathrm{c}, r}(\cdot)$ denote the function to generate the intermediate output by using client $n$'s client-side model in the $r$-th training round.
We have $\mathbf{z}_{n}^{r} = \Phi_{n}^{\mathrm{c}, r}(\mathbf{w}_{n, \boldsymbol{\lambda}^{r}}^{\mathrm{c}, r}, \mathcal{D}_{n})$.

When a client has finished its client-side model FP, it transmits its intermediate output and the corresponding labels to the central server via wireless channels.
Let $\mathcal{K} = \{1, 2, \ldots, K\}$ denote the set of $K$ RBs.
We denote $u_{n,k}^{r}\in\{0,1\}$ as the allocation decision of the $k$-th RB for client $n \in \mathcal{N}$ in the $r$-th training round.
If $u_{n,k}^{r} = 1$, then the $k$-th RB is allocated to client $n$.
Otherwise, $u_{n,k}^{r} = 0$.
Each RB is allocated to only one client.
That is, $\sum_{n=1}^{N}u_{n,k}^{r} \leq 1, k \in \mathcal{K}, r\in\mathcal{R}$.
Let $ \mathbf{U}^{r} \in \mathbb{R}^{N \times K}$ denote the RB allocation decision matrix in the $r$-th training round, with $\mathbf{u}_{n}^{r} = (u_{n,1}^{r}, u_{n,2}^{r}, \ldots, u_{n,K}^{r})^{\intercal} \in \mathbb{R}^{1 \times N}$ denoting the $n$-th row vector.
We denote $p_{n}^{r} \in [0, P^{\max}]$ as client $n$'s uplink transmit power in the $r$-th training round, where $P^{\max}$ denotes the maximum transmit power.
Let $\mathbf{p}^{r} = (p_{1}^{r}, p_{2}^{r}, \ldots, p_{N}^{r}) \in \mathbb{R}^{N}$ denote the transmit power decision vector for all clients in the $r$-th training round.
We consider that packet errors can occur in the intermediate output transmission. 
Each client transmits the intermediate output in a single packet.
The packet error rate of the intermediate output transmission of client $n\in\mathcal{N}$ in the $r$-th training round is given by \cite{9210812}: 
\begin{align}
    \!\!\!s_{n}^{r}(\mathbf{u}_{n}^{r}, p_{n}^{r}) = \mathbb{E}_{\lvert h_{n}^{r}\rvert}\left[1-\mathrm{exp}\left(-\frac{\alpha BN_{0}\sum_{k=1}^{K} u_{n,k}^{r}}{p_{n}^{r}\lvert h_{n}^{r}\rvert^{2}}\right)\right],
\end{align}
where $\alpha$, $B$, and $N_{0}$ denote the waterfall threshold \cite{5703199}, the bandwidth of an RB, and the received noise power spectral density, respectively.
$\lvert h_{n}^{r}\rvert$ denotes the channel gain between client $n$ and the central server in the $r$-th training round. 

Let $\hat{\mathbf{z}}_{n}^{r}$ denote client $n$'s intermediate output, which will be used by the central server for further training with client $n$'s server-side model in the $r$-th training round.
Let $\beta_{n}^{r}$ denote the binary indicator of packet error experienced by client $n$ in the $r$-th training round.
If the received intermediate output has no packet error (with probability $1-s_{n}^{r}(\mathbf{u}_{n}^{r}, p_{n}^{r})$), then $\beta_{n}^{r}$ is equal to 1.
Otherwise, $\beta_{n}^{r}$ is equal to 0.
That is, we have $\mathbb{P}(\beta_{n}^{r} = 1) = 1 - s_{n}^{r}(\mathbf{u}_{n}^{r}, p_{n}^{r})$ and $\mathbb{P}(\beta_{n}^{r} = 0) = s_{n}^{r}(\mathbf{u}_{n}^{r}, p_{n}^{r})$, respectively.
We denote $\boldsymbol{\beta}^{r} = (\beta_{1}^{r}, \beta_{2}^{r}, \ldots, \beta_{N}^{r}) \in \mathbb{R}^{N}$ as the indicator vector of all clients.
Similar to \cite{9210812}, the central server discards packets with errors.
Thus, $\hat{\mathbf{z}}_{n}^{r}$ is given by
$\hat{\mathbf{z}}_{n}^{r} = \beta_{n}^{r}\mathbf{z}_{n}^{r}$.

\textbf{Stage 3}:  
The central server performs FP.
Let $\Phi_{n}^{\mathrm{s}, r}(\cdot)$ denote the function to generate the FP output by using client $n$'s server-side model in the $r$-th training round.
The generated output is given by
$\hat{\mathbf{y}}_{n}^{r} = \Phi_{n}^{\mathrm{s}, r}(\mathbf{w}_{n, \boldsymbol{\lambda}^{r}}^{\mathrm{s}, r}, \hat{\mathbf{z}}_{n}^{r})$.

The central server performs BP over client $n$'s server-side model to minimize the loss $F_{n}^{r}(\mathbf{w}_{n}^{r}; \bm{\xi}_{n}) = L_{F}(\hat{\mathbf{y}}_{n}^{r}, \mathbf{y}_{n})$,
where $L_{F}(\cdot)$ denotes the loss function (e.g., cross-entropy loss).
Let $\tilde{\mathcal{N}}^{r}$ and $\nabla F_{n}^{\mathrm{s}, r}(\mathbf{w}_{n, \boldsymbol{\lambda}^{r}}^{\mathrm{s}, r}; \bm{\xi}_{n})$ denote the set of clients with no packet errors in the received intermediate output and the gradient of client $n$'s server-side model in the $r$-th training round, respectively.
In the $r$-th training round, client $n$'s server-side model is updated as
\begin{equation}
\label{eq:server_model_update}
    \mathbf{w}_{n}^{\mathrm{s}, r+1} = \mathbf{w}_{n, \boldsymbol{\lambda}^{r}}^{\mathrm{s}, r} - \eta\nabla F_{n}^{\mathrm{s}, r}(\mathbf{w}_{n, \boldsymbol{\lambda}^{r}}^{\mathrm{s}, r}; \bm{\xi}_{n}), \quad n\in\tilde{\mathcal{N}}^{r},
\end{equation}
where $\eta$ denotes the learning rate.
When all server-side models have been updated, the central server performs aggregation as
$\bar{\mathbf{w}}^{\mathrm{s}, r+1} = \sum_{n\in\tilde{\mathcal{N}}^{r}} D_{n}\mathbf{w}_{n}^{\mathrm{s}, r+1}/\sum_{j\in\tilde{\mathcal{N}}^{r}} D_{j}$,
The central server then sends the first layer of gradient $\nabla F_{n}^{\mathrm{s}, r}(\mathbf{w}_{n, \boldsymbol{\lambda}^{r}}^{\mathrm{s}, r}; \bm{\xi}_{n})$ (i.e., the gradient of the $(\Vert \boldsymbol{\lambda}^{r}\Vert_{0} + 1)$-th layer of $\mathbf{w}_{n}^{r}$) back to client $n$.
Client $n$ performs BP by using the received gradient. 
Let $\nabla F_{n}^{\mathrm{c}, r}(\mathbf{w}_{n}^{r}; \bm{\xi}_{n})$ denote the gradient of client $n$'s client-side model in the $r$-th training round.
In the $r$-th training round, client $n$ updates its client-side model as
\begin{equation}
\label{eq:client_model_update}
    \mathbf{w}_{n}^{\mathrm{c}, r+1} = \mathbf{w}_{n, \boldsymbol{\lambda}^{r}}^{\mathrm{c}, r} - \eta\nabla F_{n}^{\mathrm{c}, r}(\mathbf{w}_{n, \boldsymbol{\lambda}^{r}}^{\mathrm{c}, r}; \bm{\xi}_{n}), \quad n\in\tilde{\mathcal{N}}^{r}.
\end{equation}

Since the model update of ASFL depends on decision variables $\{\mathbf{U}^{r}, \mathbf{p}^{r},  \boldsymbol{\lambda}^{r}\}_{r\in\mathcal{R}}$, we characterize their impact on the convergence rate in Section \ref{sec:theoretical_analysis}.

\subsection{Delay and Energy Consumption Model}
In this subsection, we present the delay and energy consumption in each training round as functions of decision variables $\{\mathbf{U}^{r}, \mathbf{p}^{r},  \boldsymbol{\lambda}^{r}\}_{r\in\mathcal{R}}$.
In the $r$-th training round, since $\boldsymbol{\lambda}^{r-1}$ has been determined, it is not a decision variable. 

\subsubsection{Delay}
\begin{figure}[t]
    \centering
    \includegraphics[width=0.48\textwidth]{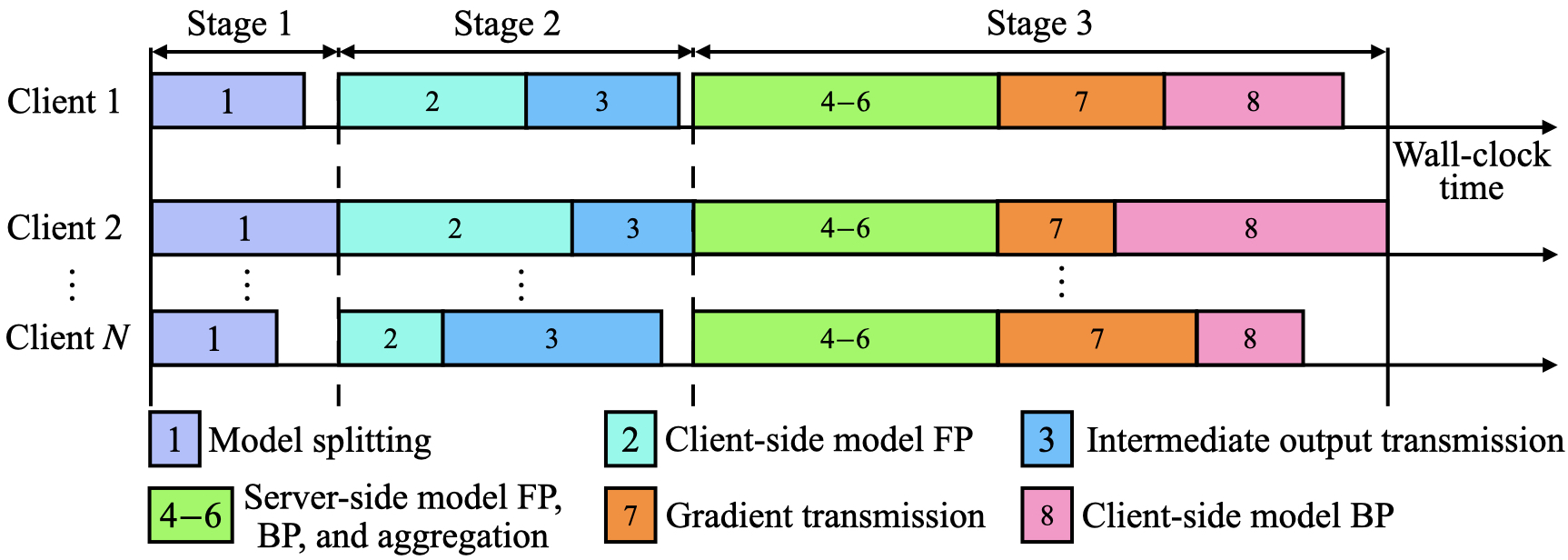}
    \caption{An illustration of the delay in one ASFL training round.}
    \label{fig:diagram}
\end{figure}

Similar to other works on SFL (e.g., \cite{lin2023efficient}), we consider synchronous learning where all clients begin each stage simultaneously.
An illustration of the delay in one ASFL training round is shown in Fig. \ref{fig:diagram}.
We now present the delay incurred in each stage.

In stage 1, the delay is incurred due to adaptive model splitting.
That is, the clients or the central server may need to send several middle layers to the other based on the model splitting decisions.
We use orthogonal frequency division multiple access (OFDMA) for transmission.
In the $r$-th training round, for client $n\in\mathcal{N}$, the downlink transmission rate $c_{n}^{\mathrm{DN}, r}$ and the uplink transmission rate $c_{n}^{\mathrm{UP}, r}$ are as follows:
\begin{align}
    &c_{n}^{\mathrm{DN}, r} = B^{\mathrm{DN}}\log_{2}\left(1 + \frac{p^{\mathrm{s}}\lvert h_{n}^{r}\rvert^{2}}{B^{\mathrm{DN}}N_{0}}\right), \\
    &c_{n}^{\mathrm{UP}, r}(\mathbf{u}_{n}^{r}, p_{n}^{r}) = \sum_{k=1}^{K}u_{n,k}^{r}B\log_{2}\left(1 + \frac{p_{n}^{r}\lvert h_{n}^{r}\rvert^{2}}{BN_{0}}\right), 
\end{align}
where $B^{\mathrm{DN}}$ and $p^{\mathrm{s}}$ denote the downlink bandwidth and the transmit power of the central server, respectively.
Let $T_{n}^{\mathrm{S1}, r}$ denote the delay of stage 1 of client $n$ in the $r$-th training round.
Let $\boldsymbol{\psi} = (\psi_{1}, \psi_{2}, \ldots, \psi_{M}) \in\mathbb{R}^{M}$ denote the model size vector, with $\psi_{m}$ representing the size of the $m$-th layer (in bits).
If $\Vert\boldsymbol{\lambda}^{r}\Vert_{0} \geq \Vert\boldsymbol{\lambda}^{r-1}\Vert_{0}$, then the size of the transmitted model parameters of the middle layers is equal to $(\boldsymbol{\lambda}^{r} - \boldsymbol{\lambda}^{r-1})^{\intercal}\boldsymbol{\psi}$.
Otherwise, the size of the transmitted model parameters of the middle layers is $(\boldsymbol{\lambda}^{r-1} - \boldsymbol{\lambda}^{r})^{\intercal}\boldsymbol{\psi}$.
Hence, $T_{n}^{\mathrm{S1}, r}$ satisfies
\begin{align}
    T_{n}^{\mathrm{S1}, r}(\mathbf{u}_{n}^{r}, p_{n}^{r}, \boldsymbol{\lambda}^{r}) = 
    \begin{cases}
        \frac{(\boldsymbol{\lambda}^{r} - \boldsymbol{\lambda}^{r-1})^{\intercal}\boldsymbol{\psi}}{c_{n}^{\mathrm{DN}, r}}, &  \text{if $\Vert\boldsymbol{\lambda}^{r}\Vert_{0} \geq \Vert\boldsymbol{\lambda}^{r-1}\Vert_{0}$},  \\
        \frac{(\boldsymbol{\lambda}^{r-1} - \boldsymbol{\lambda}^{r})^{\intercal}\boldsymbol{\psi}}{c_{n}^{\mathrm{UP}, r}(\mathbf{u}_{n}^{r}, p_{n}^{r})}, &  \text{otherwise}.  
    \end{cases}
\end{align}
The delay of stage 1 in the $r$-th training round is given by
\begin{align}
    T^{\mathrm{S1}, r}(\mathbf{U}^{r}, \mathbf{p}^{r}, \boldsymbol{\lambda}^{r}) = \max_{n\in\mathcal{N}}\{T_{n}^{\mathrm{S1}, r}(\mathbf{u}_{n}^{r}, p_{n}^{r}, \boldsymbol{\lambda}^{r})\}.
\end{align}

Stage 2 has two processes: the client-side model FP and the intermediate output transmission.
Let $\mathbf{v}^{\mathrm{FP}} = (v_{1}^{\mathrm{FP}}, \ldots, v_{M}^{\mathrm{FP}}) \in \mathbb{R}^{M}$ denote the vector of computation workload for FP, with $v_{m}^{\mathrm{FP}}$ denoting the computation workload for FP of the $m$-th layer (in floating point operations (FLOPs)).
The delay of the client-side model FP of client $n$ in the $r$-th training round is as follows:
\begin{align}
    T_{n}^{\mathrm{CFP}, r}(\boldsymbol{\lambda}^{r}) = \frac{\kappa^{\mathrm{c}}(\boldsymbol{\lambda}^{r})^{\intercal}\mathbf{v}^{\mathrm{FP}}D_{n}}{f_{n}},
\end{align}
where $f_{n}$ and $\kappa^{\mathrm{c}}$ denote client $n$'s central processing unit (CPU) frequency and the number of CPU cycles required for a client to complete one FLOP, respectively.
Let $q_{m}$ denote the output size of the $m$-th layer (in bits).
Since $\boldsymbol{\lambda}^{r}$ is a binary vector and $\lambda_{m}^{r} \leq \lambda_{m-1}^{r},  m \in \mathcal{M}\backslash\{1\}$, the size of the intermediate output at the $\Vert \boldsymbol{\lambda}^{r}\Vert$-th layer can be represented by $\sum_{m=1}^{M-1}(\lambda_{m}^{r} - \lambda_{m+1}^{r})q_{m}$.
The delay of the intermediate output transmission of client $n$ in the $r$-th training round is
\begin{align}
    T_{n}^{\mathrm{UP}, r}(\mathbf{u}_{n}^{r}, p_{n}^{r}, \boldsymbol{\lambda}^{r}) = \frac{D_{n}\sum_{m=1}^{M-1}(\lambda_{m}^{r} - \lambda_{m+1}^{r})q_{m}}{c_{n}^{\mathrm{UP}, r}(\mathbf{u}_{n}^{r}, p_{n}^{r})}.
\end{align}
The total delay of stage 2 in the $r$-th training round is
\begin{align}
    &T^{\mathrm{S2}, r}(\mathbf{U}^{r}, \mathbf{p}^{r}, \boldsymbol{\lambda}^{r}) \nonumber\\
    =&\: \max_{n\in\mathcal{N}}\{T_{n}^{\mathrm{CFP}, r}(\boldsymbol{\lambda}^{r}) + T_{n}^{\mathrm{UP}, r}(\mathbf{u}_{n}^{r}, p_{n}^{r}, \boldsymbol{\lambda}^{r})\}.
\end{align}
    
Stage 3 has four processes: the server-side model FP, the server-side model BP, the gradient transmission, and the client-side model BP.
Since $\beta_{n}^{r}$ is a random variable, we define the expected delay of the server-side model FP in the $r$-th training round as the time required for the central server to complete the FLOPs of all server-side model FP, which is given by
\begin{align}
    &\mathbb{E}_{\boldsymbol{\beta}^{r}}[T^{\mathrm{SFP}, r}(\mathbf{U}^{r}, \mathbf{p}^{r}, \boldsymbol{\lambda}^{r})] \nonumber\\
    =&\: \frac{\kappa^{\mathrm{s}}(\mathbf{1}_{M}-\boldsymbol{\lambda}^{r})^{\intercal}\mathbf{v}^{\mathrm{FP}}\sum_{n=1}^{N}\big(1-s_{n}^{r}(\mathbf{u}_{n}^{r}, p_{n}^{r})\big)D_{n}}{f^{\mathrm{s}}},
\end{align}
where $f^{\mathrm{s}}$ and $\kappa^{\mathrm{s}}$ denote the CPU frequency of the central server and the number of CPU cycles required for the central server to complete one FLOP, respectively.
The expected delay of the server-side model BP in the $r$-th training round is
\begin{align}
    &\mathbb{E}_{\boldsymbol{\beta}^{r}}[T^{\mathrm{SBP}, r}(\mathbf{U}^{r}, \mathbf{p}^{r}, \boldsymbol{\lambda}^{r})] \nonumber\\
    =&\:         \frac{\kappa^{\mathrm{s}}(\mathbf{1}_{M}-\boldsymbol{\lambda}^{r})^{\intercal}\mathbf{v}^{\mathrm{BP}}\sum_{n=1}^{N}\big(1-s_{n}^{r}(\mathbf{u}_{n}^{r}, p_{n}^{r})\big)D_{n}}{f^{\mathrm{s}}},
\end{align}
where $\mathbf{v}^{\mathrm{BP}} = (v_{1}^{\mathrm{BP}}, \ldots, v_{M}^{\mathrm{BP}}) \in \mathbb{R}^{M}$ denotes the vector of computation workload for BP, with $v_{m}^{\mathrm{BP}}$ denoting the computation workload for BP of the $m$-th layer (in FLOPs).
The size of the gradient of the $(\Vert \boldsymbol{\lambda}^{r}\Vert_{0} + 1)$-th layer can be represented by $\sum_{m=1}^{M-1}(\lambda_{m}^{r} - \lambda_{m+1}^{r})\psi_{m+1}$.
The expected delay for the gradient transmission to client $n\in\mathcal{N}$ in the $r$-th training round is 
\begin{align}
    &\mathbb{E}_{\beta_{n}^{r}}[T_{n}^{\mathrm{DN}, r}(\mathbf{u}_{n}^{r}, p_{n}^{r}, \boldsymbol{\lambda}^{r})] 
     \nonumber\\
    =&\:\frac{\big(1\!-\!s_{n}^{r}(\mathbf{u}_{n}^{r}, p_{n}^{r})\big)D_{n}\sum_{m=1}^{M-1}(\lambda_{m}^{r} - \lambda_{m+1}^{r})\psi_{m+1}}{c_{n}^{\mathrm{DN}, r}}.
\end{align}
The expected delay of the client-side model BP of client $n\in\mathcal{N}$ in the $r$-th training round is obtained as
\begin{align}
    \!\mathbb{E}_{\beta_{n}^{r}}[T_{n}^{\mathrm{CBP}, r}(\mathbf{u}_{n}^{r}, p_{n}^{r}, \boldsymbol{\lambda}^{r})] 
    \!=\! \frac{\kappa^{\mathrm{c}}\big(1\!-\!s_{n}^{r}(\mathbf{u}_{n}^{r}, p_{n}^{r})\big)(\boldsymbol{\lambda}^{r})^{\intercal}\mathbf{v}^{\mathrm{BP}}D_{n}}{f_{n}}.
\end{align}
The expected total delay of all clients in stage 3 of the $r$-th training round is given by
\begin{align}
    &\mathbb{E}_{\boldsymbol{\beta}^{r}}[T^{\mathrm{S3}, r}(\mathbf{U}^{r}, \mathbf{p}^{r}, \boldsymbol{\lambda}^{r})] \nonumber\\
    =&\: \max_{n\in\mathcal{N}}\{\mathbb{E}_{\boldsymbol{\beta}^{r}}[T^{\mathrm{SFP}, r}(\mathbf{U}^{r}, \mathbf{p}^{r}, \boldsymbol{\lambda}^{r}) + T^{\mathrm{SBP}, r}(\mathbf{U}^{r}, \mathbf{p}^{r}, \boldsymbol{\lambda}^{r}) \nonumber\\
    &+ T_{n}^{\mathrm{DN}, r}(\mathbf{u}_{n}^{r}, p_{n}^{r}, \boldsymbol{\lambda}^{r}) + T_{n}^{\mathrm{CBP}, r}(\mathbf{u}_{n}^{r}, p_{n}^{r}, \boldsymbol{\lambda}^{r})]\}.
\end{align}
Thus, the expected total delay in the $r$-th training round is
\begin{align}
    &\mathbb{E}_{\boldsymbol{\beta}^{r}}[T^{r}(\mathbf{U}^{r}, \mathbf{p}^{r}, \boldsymbol{\lambda}^{r})] 
    = T^{\mathrm{S1}, r}(\mathbf{U}^{r}, \mathbf{p}^{r}, \boldsymbol{\lambda}^{r}) \nonumber\\
    &+ T^{\mathrm{S2}, r}(\mathbf{U}^{r}, \mathbf{p}^{r}, \boldsymbol{\lambda}^{r}) + \mathbb{E}_{\boldsymbol{\beta}^{r}}[T^{\mathrm{S3}, r}(\mathbf{U}^{r}, \mathbf{p}^{r}, \boldsymbol{\lambda}^{r})].
\end{align}

\subsubsection{Energy Consumption}
Each client incurs energy consumption through four processes: model splitting, client-side model FP, intermediate output transmission, and client-side model BP.
We do not consider the energy consumption at the central server due to its continuous power supply.

The energy consumption of client $n\in \mathcal{N}$ for model splitting in the $r$-th training round is given by
\begin{align}
    E_{n}^{\mathrm{MS}, r}(\mathbf{u}_{n}^{r}, p_{n}^{r}, \boldsymbol{\lambda}^{r}) = \frac{(\boldsymbol{\lambda}^{r-1} - \boldsymbol{\lambda}^{r})^{\intercal}\boldsymbol{\psi}p_{n}^{r}}{c_{n}^{\mathrm{UP}, r}(\mathbf{u}_{n}^{r}, p_{n}^{r})}.
\end{align}
The energy consumption of client $n\in\mathcal{N}$ for client-side FP in the $r$-th training round is the energy required for client $n$ to process all training samples, which is given by
\begin{align}
    E_{n}^{\mathrm{FP}, r}(\boldsymbol{\lambda}^{r}) = D_{n}(\boldsymbol{\lambda}^{r})^{\intercal}\mathbf{v}^{\mathrm{FP}}\phi\kappa^{\mathrm{c}}(f_{n})^{2},
\end{align}
where $\phi$ is the energy consumption coefficient.
The energy consumption of client $n\in\mathcal{N}$ for the intermediate output transmission in the $r$-th training round can be expressed as
\begin{align}
    E_{n}^{\mathrm{UP}, r}(\mathbf{u}_{n}^{r}, p_{n}^{r}, \boldsymbol{\lambda}^{r}) = p_{n}^{r}T_{n}^{\mathrm{UP}, r}(\mathbf{u}_{n}^{r}, p_{n}^{r}, \boldsymbol{\lambda}^{r}).
\end{align}
The expected energy consumption of client $n\in\mathcal{N}$ for client-side model BP in the $r$-th training round is given by
\begin{align}
    &\mathbb{E}_{\beta_{n}^{r}}[E_{n}^{\mathrm{CBP}, r}(\mathbf{u}_{n}^{r}, p_{n}^{r}, \boldsymbol{\lambda}^{r})] \nonumber\\
    =&\: \big(1-s_{n}^{r}(p_{n}^{r}, \mathbf{u}_{n}^{r})\big)D_{n}(\boldsymbol{\lambda}^{r})^{\intercal}\mathbf{v}^{\mathrm{BP}}\phi\kappa^{\mathrm{c}}(f_{n})^{2}.
\end{align}
The expected total energy consumption of client $n$ in the $r$-th training round is obtained as
\begin{align}
    &\mathbb{E}_{\beta_{n}^{r}}[E_{n}^{r}(\mathbf{u}_{n}^{r}, p_{n}^{r}, \boldsymbol{\lambda}^{r})] 
    = E_{n}^{\mathrm{MS}, r}(\mathbf{u}_{n}^{r}, p_{n}^{r}, \boldsymbol{\lambda}^{r}) + E_{n}^{\mathrm{FP}, r}(\boldsymbol{\lambda}^{r}) \nonumber\\
    &+ E_{n}^{\mathrm{UP}, r}(\mathbf{u}_{n}^{r}, p_{n}^{r}, \boldsymbol{\lambda}^{r}) 
    + \mathbb{E}_{\beta_{n}^{r}}[E_{n}^{\mathrm{CBP}, r}(\mathbf{u}_{n}^{r}, p_{n}^{r}, \boldsymbol{\lambda}^{r})].
\end{align}

\section{Theoretical Analysis and Problem Formulation}
\label{sec:theoretical_analysis}
In this section, we first characterize the impact of decision variables on the convergence rate of our proposed ASFL framework.
Then, we present the problem formulation.

\subsection{Theoretical Analysis}
Without loss of generality, we conduct analysis under non-convex loss functions.
We first present the following assumptions which are widely used in the literature (e.g., \cite{li2019convergence, WangLLJP20, 9261995}). 

\noindent\textbf{Assumption 1.} 
\textit{The loss function of each client $n \in \mathcal{N}$ is continuously differentiable and $L$-smooth.
That is, for arbitrary two vectors $\mathbf{w}_{n}^{r}$ and $\tilde{\mathbf{w}}_{n}^{r}$, we have}
$F_{n}(\mathbf{w}_{n}^{r}) \leq F_{n}(\tilde{\mathbf{w}}_{n}^{r}) + \langle \nabla F_{n}(\tilde{\mathbf{w}}_{n}^{r}), \mathbf{w}_{n}^{r} - \tilde{\mathbf{w}}_{n}^{r}\rangle + \frac{L}{2}\Vert\mathbf{w}_{n}^{r} - \tilde{\mathbf{w}}_{n}^{r}\Vert^{2}.$


\noindent\textbf{Assumption 2.} 
\textit{The variance of the local stochastic gradient of each client $n \in \mathcal{N}$ is upper-bounded, i.e., } $\mathbb{E}_{\bm{\xi}_{n}\sim\mathcal{D}_{n}} \big[\Vert\nabla F_{n}(\mathbf{w}_{n}^{r}; \bm{\xi}_{n}) - \nabla F_{n}(\mathbf{w}_{n}^{r}) \Vert^{2}\big] \leq \sigma_{F}^{2}$.

\noindent\textbf{Assumption 3.} 
\textit{The expected square norm of the gradient of each client $n \in \mathcal{N}$ is upper-bounded, i.e., } $\mathbb{E}_{\bm{\xi}_{n}\sim\mathcal{D}_{n}}[\Vert\nabla F_{n}(\mathbf{w}_{n}^{r}; \bm{\xi}_{n})\Vert^{2}] \leq \psi_{F}^{2}$.

We introduce two lemmas which are widely used in the literature (e.g., \cite{li2019convergence, WangLLJP20}) to facilitate our proof.

\noindent\textbf{Lemma 1.} \textit{For arbitrary vector $\mathbf{w}_{n}^{r}$, $n\in\mathcal{N}$, we have $\Vert\sum_{n=1}^{N}\mathbf{w}_{n}^{r}\Vert^{2} \leq N\sum_{n=1}^{N}\Vert\mathbf{w}_{n}^{r}\Vert^{2}$.}

\noindent\textbf{Lemma 2.} \textit{For arbitrary two vectors $\mathbf{w}_{n}^{r}$ and $\tilde{\mathbf{w}}_{n}^{r}$, $n\in\mathcal{N}$, we have $\langle\mathbf{w}_{n}^{r}, \tilde{\mathbf{w}}_{n}^{r}\rangle = \frac{1}{2}(\Vert\mathbf{w}_{n}^{r}\Vert^{2} + \Vert\tilde{\mathbf{w}}_{n}^{r}\Vert^{2} - \Vert\mathbf{w}_{n}^{r}-\tilde{\mathbf{w}}_{n}^{r}\Vert^{2})$.}

We denote the average client-side model and server-side model after model splitting at the beginning of the $r$-th training round as $\bar{\mathbf{w}}_{\boldsymbol{\lambda}^{r}}^{\mathrm{c}, r}$ and $\bar{\mathbf{w}}_{\boldsymbol{\lambda}^{r}}^{\mathrm{s}, r}$, respectively.
Now, we present the convergence rate of our proposed ASFL in Theorem 1.

\noindent\textbf{Theorem 1.} 
\textit{
Under Assumptions 1\:$-$\:3 and Lemmas 1\:$-$\:2, 
the convergence rate of our proposed ASFL is bounded by}
\begin{align}
\label{eq:APSL_final_result}
    &\frac{1}{R}\sum_{r=1}^{R}\Vert\nabla F(\bar{\mathbf{w}}^{r})\Vert^{2}
    \leq \frac{2}{ R}\big(F(\bar{\mathbf{w}}^{1}) - F(\mathbf{w}^{\star})\big) + 3(\sigma_{F}^{2} + \psi_{F}^{2}) \nonumber\\
    +&\: \underbrace{\frac{2(L+2)}{NR}\sum_{r=1}^{R}\sum_{n=1}^{N}(\Vert\bar{\mathbf{w}}_{\boldsymbol{\lambda}^{r}}^{\mathrm{c}, r} - \mathbf{w}_{n, \boldsymbol{\lambda}^{r}}^{\mathrm{c}, r}\Vert^{2} + (1 - \beta_{n}^{r})^{2}\Vert\bar{\mathbf{w}}_{\boldsymbol{\lambda}^{r}}^{\mathrm{s}, r}\Vert^{2})}_\text{Average long-term model discrepancies} \nonumber\\
    +&\: 2(L+2)\eta^{2}\psi_{F}^{2}. 
\end{align}

\begin{proof}
The model of client $n$ at the beginning of the $(r+1)$-th training round is given by $\mathbf{w}_{n}^{r+1} = \mathbf{w}_{n}^{r} - \beta_{n}^{r}\eta\nabla F_{n}(\mathbf{w}_{n}^{r}; \bm{\xi}_{n}^{r})$.
We can derive $\bar{\mathbf{w}}^{r+1}$ as follows:
\begin{align}
\label{eq:avg_model}
    \bar{\mathbf{w}}^{r+1} 
    =\: &\Big[\frac{1}{N}\sum_{n=1}^{N}\big(\mathbf{w}_{n, \boldsymbol{\lambda}^{r}}^{\mathrm{c}, r} -\beta_{n}^{r}\eta\nabla F_{n}^{\mathrm{c}}(\mathbf{w}_{n}^{r}; \bm{\xi}_{n}^{r})\big); \nonumber\\
    & \frac{1}{N}\sum_{n=1}^{N}\beta_{n}^{r}\big(\mathbf{w}_{n, \boldsymbol{\lambda}^{r}}^{\mathrm{s}, r} - \eta\nabla F_{n}^{\mathrm{s}}(\mathbf{w}_{n}^{r}; \bm{\xi}_{n}^{r})\big) \Big] \nonumber\\
    =\: &\Big[\frac{1}{N}\sum_{n=1}^{N}\mathbf{w}_{n, \boldsymbol{\lambda}^{r}}^{\mathrm{c}, r}; \frac{1}{N}\sum_{n=1}^{N}\beta_{n}^{r}\mathbf{w}_{n, \boldsymbol{\lambda}^{r}}^{\mathrm{s}, r}\Big] \nonumber\\
    &- \frac{1}{N}\sum_{n=1}^{N}\beta_{n}^{r}\eta\nabla F_{n}(\mathbf{w}_{n}^{r}; \bm{\xi}_{n}^{r}).
\end{align}
Based on eqn. (\ref{eq:avg_model}), we expand $\mathbb{E}[F(\bar{\mathbf{w}}^{r+1})]$ as follows:
\begin{align}
\label{eq:convergence_1}
    & \mathbb{E}[F(\bar{\mathbf{w}}^{r+1})] \nonumber\\
    =&\: \mathbb{E}\Big[F\Big(\bar{\mathbf{w}}^{r} - \bar{\mathbf{w}}^{r} + \Big[\frac{1}{N}\sum_{n=1}^{N}\mathbf{w}_{n, \boldsymbol{\lambda}^{r}}^{\mathrm{c}, r}; \frac{1}{N}\sum_{n=1}^{N}\beta_{n}^{r}\mathbf{w}_{n, \boldsymbol{\lambda}^{r}}^{\mathrm{s}, r}\Big] \nonumber\\
    &- \frac{\eta}{N}\sum_{n=1}^{N}\beta_{n}^{r}\nabla F_{n}(\mathbf{w}_{n}^{r}; \bm{\xi}_{n}^{r})\Big)\Big]  \nonumber\\
    \overset{(\mathrm{a})}{\leq}&\: \mathbb{E}[F(\bar{\mathbf{w}}^{r})] - \mathbb{E}\Big[\Big\langle \nabla F(\bar{\mathbf{w}}^{r}), \bar{\mathbf{w}}^{r} - \Big[\frac{1}{N}\sum_{n=1}^{N}\mathbf{w}_{n, \boldsymbol{\lambda}^{r}}^{\mathrm{c}, r}; \nonumber\\
    &\frac{1}{N}\sum_{n=1}^{N}\beta_{n}^{r}\mathbf{w}_{n, \boldsymbol{\lambda}^{r}}^{\mathrm{s}, r}\Big] + \frac{\eta}{N}\sum_{n=1}^{N}\beta_{n}^{r}\nabla F_{n}(\mathbf{w}_{n}^{r}; \bm{\xi}_{n}^{r})\Big\rangle\Big]  \nonumber\\
    &+ \frac{L}{2}\mathbb{E}\Big[\Big\Vert \Big[\frac{1}{N}\sum_{n=1}^{N}\mathbf{w}_{n, \boldsymbol{\lambda}^{r}}^{\mathrm{c}, r}; \frac{1}{N}\sum_{n=1}^{N}\beta_{n}^{r}\mathbf{w}_{n, \boldsymbol{\lambda}^{r}}^{\mathrm{s}, r}\Big] - \bar{\mathbf{w}}^{r} \nonumber\\
    &- \frac{\eta}{N}\sum_{n=1}^{N}\beta_{n}^{r}\nabla F_{n}(\mathbf{w}_{n}^{r}; \bm{\xi}_{n}^{r})\Big\Vert^{2}\Big]  \nonumber\\
    \overset{(\mathrm{b})}{=}&\: \mathbb{E}[F(\bar{\mathbf{w}}^{r})] - \frac{1}{2}\Big(\mathbb{E}\Big[\Vert\nabla F(\bar{\mathbf{w}}^{r})\Vert^{2} +  \Big\Vert \bar{\mathbf{w}}^{r} - \Big[\frac{1}{N}\sum_{n=1}^{N}\mathbf{w}_{n, \boldsymbol{\lambda}^{r}}^{\mathrm{c}, r}; \nonumber\\
    &\frac{1}{N}\sum_{n=1}^{N}\beta_{n}^{r}\mathbf{w}_{n, \boldsymbol{\lambda}^{r}}^{\mathrm{s}, r}\Big] + \frac{\eta}{N}\sum_{n=1}^{N}\beta_{n}^{r}\nabla F_{n}(\mathbf{w}_{n}^{r}; \bm{\xi}_{n}^{r})\Big\Vert^{2}\Big]  \nonumber\\
    &- \mathbb{E}\Big[\Big\Vert\nabla F(\bar{\mathbf{w}}^{r}) - \bar{\mathbf{w}}^{r} + \Big[\frac{1}{N}\sum_{n=1}^{N}\mathbf{w}_{n, \boldsymbol{\lambda}^{r}}^{\mathrm{c}, r}; \frac{1}{N}\sum_{n=1}^{N}\beta_{n}^{r}\mathbf{w}_{n, \boldsymbol{\lambda}^{r}}^{\mathrm{s}, r}\Big] \nonumber\\
    &- \frac{\eta}{N}\sum_{n=1}^{N}\beta_{n}^{r}\nabla F_{n}(\mathbf{w}_{n}^{r}; \bm{\xi}_{n}^{r})\Big\Vert^{2}\Big]\Big)  + \frac{L}{2}\mathbb{E}\Big[\Big\Vert \Big[\frac{1}{N}\sum_{n=1}^{N}\mathbf{w}_{n, \boldsymbol{\lambda}^{r}}^{\mathrm{c}, r}; \nonumber\\
    &\frac{1}{N}\sum_{n=1}^{N}\beta_{n}^{r}\mathbf{w}_{n, \boldsymbol{\lambda}^{r}}^{\mathrm{s}, r}\Big] - \bar{\mathbf{w}}^{r} - \frac{\eta}{N}\sum_{n=1}^{N}\beta_{n}^{r}\nabla F_{n}(\mathbf{w}_{n}^{r}; \bm{\xi}_{n}^{r})\Big\Vert^{2}\Big],
\end{align}
where inequality (a) results from Assumption 1.
Equality (b) is obtained by using Lemma 2.
For illustration simplicity, we define $A_{1} = \mathbb{E}\Big[\Big\Vert\nabla F(\bar{\mathbf{w}}^{r}) - \bar{\mathbf{w}}^{r} + \Big[\frac{1}{N}\sum_{n=1}^{N}\mathbf{w}_{n, \boldsymbol{\lambda}^{r}}^{\mathrm{c}, r}; \frac{1}{N}\sum_{n=1}^{N}\beta_{n}^{r}\mathbf{w}_{n, \boldsymbol{\lambda}^{r}}^{\mathrm{s}, r}\Big] - \frac{\eta}{N}\sum_{n=1}^{N}\beta_{n}^{r}\nabla F_{n}(\mathbf{w}_{n}^{r}; \bm{\xi}_{n}^{r})\Big\Vert^{2}\Big]$ and $A_{2} = \mathbb{E}\Big[\Big\Vert \Big[\frac{1}{N}\sum_{n=1}^{N}\mathbf{w}_{n, \boldsymbol{\lambda}^{r}}^{\mathrm{c}, r}; \frac{1}{N}\sum_{n=1}^{N}\beta_{n}^{r}\mathbf{w}_{n, \boldsymbol{\lambda}^{r}}^{\mathrm{s}, r}\Big] - \bar{\mathbf{w}}^{r} - \frac{\eta}{N}\sum_{n=1}^{N}\beta_{n}^{r}\nabla F_{n}(\mathbf{w}_{n}^{r}; \bm{\xi}_{n}^{r})\Big\Vert^{2}\Big]$. 
$A_{1}$ satisfies
\begin{align}
    A_{1} 
    =&\: \mathbb{E}\Big[\Big\Vert\nabla F(\bar{\mathbf{w}}^{r}) - \frac{1}{N}\sum_{n=1}^{N}\nabla F_{n}(\bar{\mathbf{w}}^{r}; \xi_{n}^{r}) \nonumber\\
    &+ \frac{1}{N}\sum_{n=1}^{N}\nabla F_{n}(\bar{\mathbf{w}}^{r}; \xi_{n}^{r}) - \bar{\mathbf{w}}^{r} + \Big[\frac{1}{N}\sum_{n=1}^{N}\mathbf{w}_{n, \boldsymbol{\lambda}^{r}}^{\mathrm{c}, r}; \nonumber\\
    &\frac{1}{N}\sum_{n=1}^{N}\beta_{n}^{r}\mathbf{w}_{n, \boldsymbol{\lambda}^{r}}^{\mathrm{s}, r}\Big] - \frac{\eta}{N}\sum_{n=1}^{N}\beta_{n}^{r}\nabla F_{n}(\mathbf{w}_{n}^{r}; \bm{\xi}_{n}^{r})\Big\Vert^{2}\Big] \nonumber\\
    \overset{(\mathrm{a})}{\leq}&\: 3\mathbb{E}\Big[\Big\Vert\frac{1}{N}\sum_{n=1}^{N}\nabla F_{n}(\bar{\mathbf{w}}^{r}) - \frac{1}{N}\sum_{n=1}^{N}\nabla F_{n}(\bar{\mathbf{w}}^{r}; \xi_{n}^{r})\Big\Vert^{2}\Big] \nonumber\\
    &+ 3\mathbb{E}\Big[\Big\Vert\frac{1}{N}\sum_{n=1}^{N}\nabla F_{n}(\bar{\mathbf{w}}^{r}; \xi_{n}^{r})\Big\Vert^{2}\Big] + 3A_{2} \nonumber\\
    \overset{(\mathrm{b})}{\leq}&\: \frac{3}{N}\sum_{n=1}^{N}\mathbb{E}\Big[\Big\Vert\nabla F_{n}(\bar{\mathbf{w}}^{r}) - \nabla F_{n}(\bar{\mathbf{w}}^{r}; \xi_{n}^{r})\Big\Vert^{2}\Big] \nonumber\\
    &+ \frac{3}{N}\sum_{n=1}^{N}\mathbb{E}\Big[\Big\Vert\nabla F_{n}(\bar{\mathbf{w}}^{r}; \xi_{n}^{r})\Big\Vert^{2}\Big] + 3A_{2} \nonumber\\
    \overset{(\mathrm{c})}{\leq}&\: 3(\sigma_{F}^{2} + \psi_{F}^{2}) + 3A_{2},
\end{align}
where inequalities (a) and (b) result from Lemma 1.
Inequality (c) results from Assumptions 2 and 3.
Therefore, $\mathbb{E}[F(\bar{\mathbf{w}}^{r+1})]$ satisfies
\begin{align}
    \label{eq:convergence_2}
    \mathbb{E}[F(\bar{\mathbf{w}}^{r+1})] 
    \leq&\: \mathbb{E}[F(\bar{\mathbf{w}}^{r})] - \frac{1}{2}\Vert\nabla F(\bar{\mathbf{w}}^{r})\Vert^{2} + \frac{3}{2}(\sigma_{F}^{2} + \psi_{F}^{2}) \nonumber\\
    &+ \frac{L+2}{2}A_{2}.
\end{align}
Then, we analyze $A_{2}$.
In particular, it satisfies
\begin{align}
    A_{2} 
    =&\: \mathbb{E}\Big[\Big\Vert \frac{\eta}{N}\sum_{n=1}^{N}\beta_{n}^{r}\nabla F_{n}(\mathbf{w}_{n}^{r}; \bm{\xi}_{n}^{r}) + \Big[\frac{1}{N}\sum_{n=1}^{N}\bar{\mathbf{w}}_{\boldsymbol{\lambda}^{r}}^{\mathrm{c}, r}; \nonumber\\
    &\frac{1}{N}\sum_{n=1}^{N}\bar{\mathbf{w}}_{\boldsymbol{\lambda}^{r}}^{\mathrm{s}, r}\Big] - \Big[\frac{1}{N}\sum_{n=1}^{N}\mathbf{w}_{n, \boldsymbol{\lambda}^{r}}^{\mathrm{c}, r}; \frac{1}{N}\sum_{n=1}^{N}\beta_{n}^{r}\mathbf{w}_{n, \boldsymbol{\lambda}^{r}}^{\mathrm{s}, r}\Big]\Big\Vert^{2} \Big] \nonumber\\
    \overset{(\mathrm{a})}{\leq}&\: 2\mathbb{E}\Big[\Big\Vert \frac{\eta}{N}\sum_{n=1}^{N}\beta_{n}^{r}\nabla F_{n}(\mathbf{w}_{n}^{r}; \bm{\xi}_{n}^{r})\Big\Vert^{2}\Big] + 2\mathbb{E}\Big[\Big\Vert\Big[\frac{1}{N}\sum_{n=1}^{N}\bar{\mathbf{w}}_{\boldsymbol{\lambda}^{r}}^{\mathrm{c}, r}; \nonumber\\
    &\frac{1}{N}\sum_{n=1}^{N}\bar{\mathbf{w}}_{\boldsymbol{\lambda}^{r}}^{\mathrm{s}, r}\Big] - \Big[\frac{1}{N}\sum_{n=1}^{N}\mathbf{w}_{n, \boldsymbol{\lambda}^{r}}^{\mathrm{c}, r}; \frac{1}{N}\sum_{n=1}^{N}\beta_{n}^{r}\mathbf{w}_{n, \boldsymbol{\lambda}^{r}}^{\mathrm{s}, r}\Big]\Big\Vert^{2}\Big] \nonumber\\
    \overset{(\mathrm{b})}{\leq}&\: \frac{2\eta^{2}\psi_{F}^{2}}{N}\sum_{n=1}^{N}(\beta_{n}^{r})^{2} + \frac{2}{N}\sum_{n=1}^{N}\Big\Vert\bar{\mathbf{w}}_{\boldsymbol{\lambda}^{r}}^{\mathrm{c}, r} - \mathbf{w}_{n, \boldsymbol{\lambda}^{r}}^{\mathrm{c}, r}\Big\Vert^{2} \nonumber\\
    &+ \frac{2}{N}\sum_{n=1}^{N}\Big\Vert\bar{\mathbf{w}}_{\boldsymbol{\lambda}^{r}}^{\mathrm{s}, r} - \beta_{n}^{r}\mathbf{w}_{n, \boldsymbol{\lambda}^{r}}^{\mathrm{s}, r}\Big\Vert^{2},
\end{align}
where inequality (a) is derived using Lemma 1. 
Inequality (b) results from Lemma 1 and Assumption 3.
Therefore, $\mathbb{E}[F(\bar{\mathbf{w}}^{r+1})]$ satisfies
\begin{align}
    \label{eq:convergence_3}
    &\mathbb{E}[F(\bar{\mathbf{w}}^{r+1})] \nonumber\\
    \leq&\: \mathbb{E}[F(\bar{\mathbf{w}}^{r})] - \frac{1}{2}\Vert\nabla F(\bar{\mathbf{w}}^{r})\Vert^{2} + \frac{3}{2}(\sigma_{F}^{2} + \psi_{F}^{2}) \nonumber\\
    &+ \frac{(L+2)\eta^{2}\psi_{F}^{2}}{N}\sum_{n=1}^{N}(\beta_{n}^{r})^{2} + \frac{L+2}{N}\sum_{n=1}^{N}\Big\Vert\bar{\mathbf{w}}_{\boldsymbol{\lambda}^{r}}^{\mathrm{c}, r} - \mathbf{w}_{n, \boldsymbol{\lambda}^{r}}^{\mathrm{c}, r}\Big\Vert^{2} \nonumber\\
    &+ \frac{L+2}{N}\sum_{n=1}^{N}\Big\Vert\bar{\mathbf{w}}_{\boldsymbol{\lambda}^{r}}^{\mathrm{s}, r} - \beta_{n}^{r}\mathbf{w}_{n, \boldsymbol{\lambda}^{r}}^{\mathrm{s}, r}\Big\Vert^{2}.
\end{align}

To determine the convergence rate, we sum up both sides of inequality (\ref{eq:convergence_3}) for all $R$ training rounds and multiply both sides by $\frac{2}{R}$.
Then, we rearrange the inequality above and obtain the convergence rate of our proposed ASFL as
\begin{align}
\label{eq:SFL_final_result}
    &\frac{1}{R}\sum_{r=1}^{R}\Vert\nabla F(\bar{\mathbf{w}}^{r})\Vert^{2} \nonumber\\
    \overset{(\mathrm{a})}{\leq} &\: \frac{2}{ R}\big(F(\bar{\mathbf{w}}^{1}) - F(\mathbf{w}^{\star})\big) + \frac{2(L+2)\eta^{2}\psi_{F}^{2}}{NR}\sum_{r=1}^{R}\sum_{n=1}^{N}\big(\beta_{n}^{r}\big)^{2} \nonumber\\
    &+ \frac{2(L+2)}{NR}\sum_{r=1}^{R}\sum_{n=1}^{N}\Big\Vert\bar{\mathbf{w}}_{\boldsymbol{\lambda}^{r}}^{\mathrm{c}, r} - \mathbf{w}_{n, \boldsymbol{\lambda}^{r}}^{\mathrm{c}, r}\Big\Vert^{2} + 3(\sigma_{F}^{2} + \psi_{F}^{2}) \nonumber\\
    & + \frac{2(L+2)}{NR}\sum_{r=1}^{R}\sum_{n=1}^{N}\Big\Vert\bar{\mathbf{w}}_{\boldsymbol{\lambda}^{r}}^{\mathrm{s}, r} - \beta_{n}^{r}\mathbf{w}_{n, \boldsymbol{\lambda}^{r}}^{\mathrm{s}, r}\Big\Vert^{2} \nonumber\\
    \overset{(\mathrm{b})}{\leq} &\: \frac{2}{ R}\big(F(\bar{\mathbf{w}}^{1}) - F(\mathbf{w}^{\star})\big) + 2(L+2)\eta^{2}\psi_{F}^{2} + 3(\sigma_{F}^{2} + \psi_{F}^{2}) \nonumber\\
    &+ \frac{2(L+2)}{NR}\sum_{r=1}^{R}\sum_{n=1}^{N}\Big\Vert\bar{\mathbf{w}}_{\boldsymbol{\lambda}^{r}}^{\mathrm{c}, r} - \mathbf{w}_{n, \boldsymbol{\lambda}^{r}}^{\mathrm{c}, r}\Big\Vert^{2} \nonumber\\
    & + \frac{2(L+2)}{NR}\sum_{r=1}^{R}\sum_{n=1}^{N}(1-\beta_{n}^{r})^{2}\Big\Vert\bar{\mathbf{w}}_{\boldsymbol{\lambda}^{r}}^{\mathrm{s}, r}\Big\Vert^{2},
\end{align}
where inequality (a) is obtained by using $F(\bar{\mathbf{w}}^{R+1}) \geq F(\mathbf{w}^{\star})$.
Inequality (b) results from the fact that $\beta_{n}^{r} \leq 1$ and $\bar{\mathbf{w}}_{\boldsymbol{\lambda}^{r}}^{\mathrm{s}, r} = \mathbf{w}_{n, \boldsymbol{\lambda}^{r}}^{\mathrm{s}, r}$.
This completes the proof of Theorem 1.    
\end{proof}

\noindent\textbf{Remark 1.} 
Theorem 1 suggests that the first two terms and the fourth term on the right-hand side of inequality (\ref{eq:APSL_final_result}) are independent of the decision variables $\{\mathbf{U}^{r}, \mathbf{p}^{r}, \boldsymbol{\lambda}^{r}\}_{r\in\mathcal{R}}$.
Thus, we should optimize the decision variables $\{\mathbf{U}^{r}, \mathbf{p}^{r}, \boldsymbol{\lambda}^{r}\}_{r\in\mathcal{R}}$ in order to minimize the average long-term model discrepancies (i.e., the third term on the right-hand side of inequality (\ref{eq:APSL_final_result})).
Recall from Section \ref{sec:learning_model} that $\beta_{n}^{r}$ is a binary random variable, which is equal to 1 with probability $1-s_{n}^{r}(\mathbf{u}_{n}^{r}, p_{n}^{r})$.
Hence, we aim to minimize the expected average long-term model discrepancies by considering the randomness of $\beta_{n}^{r}$. 
On the other hand, the decision variables $\{\mathbf{U}^{r}, \mathbf{p}^{r}, \boldsymbol{\lambda}^{r}\}_{r\in\mathcal{R}}$ can affect the delay and energy consumption significantly.
Thus, we need to determine them properly to jointly improve the learning performance and efficiency of our proposed ASFL.

\subsection{Problem Formulation}
We aim to minimize the expected average long-term model discrepancies while guaranteeing the delay and energy consumption constraints.
We formulate the problem as follows:
\begin{subequations}
\label{opt_problem_v1}
\begin{alignat}{5}
&\displaystyle{\minimize_{\substack{\{\mathbf{U}^{r}, \mathbf{p}^{r},\\ \boldsymbol{\lambda}^{r}\}_{r\in\mathcal{R}}}}} & & \quad
\frac{1}{NR}\sum_{r=1}^{R}\sum_{n=1}^{N}\Big(\Big\Vert\bar{\mathbf{w}}_{\boldsymbol{\lambda}^{r}}^{\mathrm{c}, r} - \mathbf{w}_{n, \boldsymbol{\lambda}^{r}}^{\mathrm{c}, r}\Big\Vert^{2} \nonumber\\
& & & \quad  + \big(1-s_{n}^{r}(\mathbf{u}_{n}^{r}, p_{n}^{r})\big)^{2}\Big\Vert\bar{\mathbf{w}}_{\boldsymbol{\lambda}^{r}}^{\mathrm{s}, r} \Big\Vert^{2}\Big)
 \label{eq:optProb} \\
&\mathrm{subject\: to}
& & \quad \frac{1}{R}\sum\nolimits_{r=1}^{R}\mathbb{E}_{\boldsymbol{\beta}^{r}}[T^{r}(\mathbf{U}^{r}, \mathbf{p}^{r}, \boldsymbol{\lambda}^{r})] \leq \gamma, \label{c1}\\
& & & \quad \frac{1}{R}\sum\nolimits_{r=1}^{R}\mathbb{E}_{\beta_{n}^{r}}[E_{n}^{r}(\mathbf{u}_{n}^{r}, p_{n}^{r}, \boldsymbol{\lambda}^{r})] \leq \delta, \quad n \in \mathcal{N}, \label{c9}\\
& & &\quad \sum\nolimits_{n=1}^{N}u_{n,k}^{r} \leq 1, \quad k \in \mathcal{K}, r \in \mathcal{R}, \label{c4}\\
& & &\quad 0 \leq p_{n}^{r} \leq P^{\mathrm{max}}, \quad n \in \mathcal{N}, r \in \mathcal{R}, \label{c5}\\
& & &\quad \lambda_{1}^{r} = 1, \quad r \in \mathcal{R}, \label{c8} \\
& & &\quad \lambda_{m}^{r} \leq \lambda_{m-1}^{r}, \quad m \in \mathcal{M}\backslash\{1\}, r \in \mathcal{R}, \label{c7} \\
& & &\quad \lambda_{m}^{r} \in \{0, 1\}, \quad  m \in \mathcal{M}\backslash\{1\}, r \in \mathcal{R}, \label{c2}\\
& & &\quad u_{n,k}^{r} \in \{0, 1\}, \quad n \in \mathcal{N}, k \in \mathcal{K}, r \in \mathcal{R}. \label{c6}
\end{alignat}
\end{subequations}
Constraints (\ref{c1}) and (\ref{c9}) ensure that the average delay and energy consumption on training are bounded.
Constraints (\ref{c4})\:$-$\:(\ref{c6}) are constraints for the RB allocation, transmit power allocation, and model splitting decisions.

The challenges of solving problem (\ref{opt_problem_v1}) are twofold.
First, due to the dynamic wireless channel conditions and time-varying model parameters, it is difficult to guarantee constraints (\ref{c1}) and (\ref{c9}) while minimizing the nonconvex objective function (\ref{eq:optProb}).
Second, the coupling of model splitting and resource allocation decisions complicates the solution.
To address these challenges, we propose an online optimization enhanced block coordinate descent (OOE-BCD) algorithm.

\section{OOE-BCD Algorithm}
\label{sec:methodology}
To address the aforementioned challenges, we decompose problem (\ref{opt_problem_v1}) into three subproblems and solve them iteratively in each training round.
In each subproblem, we optimize a decision variable by fixing the other two decision variables.
Our proposed OOE-BCD algorithm is presented at the end of this section.
To avoid the communication overhead of sending the entire client-side model to the central server for determining the objective value (\ref{eq:optProb}), each client $n$ randomly samples a subset of its client-side model parameters with a sampling ratio $\iota$\footnote{As in \cite{10971879}, a small value of sampling ratio is sufficient to characterize the average model discrepancies.
The corresponding communication cost can be considered to be negligible.} as $\tilde{\mathbf{w}}_{n, \boldsymbol{\lambda}^{r}}^{\mathrm{c}, r}$ and transmits the sampled client-side model to the central server. 
The average sampled client-side model is given by
$\tilde{\mathbf{w}}_{\boldsymbol{\lambda}^{r}}^{\mathrm{avg}, r} = \frac{1}{N}\sum_{n=1}^{N} \tilde{\mathbf{w}}_{n, \boldsymbol{\lambda}^{r}}^{\mathrm{c}, r}$.
The objective function (\ref{eq:optProb}) can be written as
$\frac{1}{R}\sum_{r=1}^{R}g_{\mathrm{obj}}^{r}(\mathbf{U}^{r}, \mathbf{p}^{r}, \boldsymbol{\lambda}^{r}) \approx \frac{1}{ NR}\sum_{r=1}^{R}\sum_{n=1}^{N}\big(\frac{1}{\iota}\Vert\tilde{\mathbf{w}}_{\boldsymbol{\lambda}^{r}}^{\mathrm{avg}, r} - \tilde{\mathbf{w}}_{n, \boldsymbol{\lambda}^{r}}^{\mathrm{c}, r}\Vert^{2} + (1-s_{n}^{r}(\mathbf{u}_{n}^{r}, p_{n}^{r}))^{2}\Vert\bar{\mathbf{w}}_{\boldsymbol{\lambda}^{r}}^{\mathrm{s}, r}\Vert^{2}\big)$,
where $g_{\mathrm{obj}}^{r}(\mathbf{U}^{r}, \mathbf{p}^{r}, \boldsymbol{\lambda}^{r})$ denotes the objective function in the $r$-th training round.

\subsection{Model Splitting Subproblem}
We first fix the decision variables $\{\mathbf{U}^{r}, \mathbf{p}^{r}\}_{r\in\mathcal{R}}$ and optimize the model splitting decisions $\{\boldsymbol{\lambda}^{r}\}_{r\in\mathcal{R}}$. 
The objective is denoted as $g_{\mathrm{obj}}^{r}(\boldsymbol{\lambda}^{r})$.
The challenges of solving this subproblem are twofold.
First, the model splitting decisions are coupled between two consecutive training rounds. 
When we determine $\boldsymbol{\lambda}^{r+1}$, the objective function (\ref{eq:optProb}) and constraints (\ref{c1}) and (\ref{c9}) depend on the previous decision $\boldsymbol{\lambda}^{r}$.
Second, we need to balance the long-term objective and constraints.
Hence, to address these two challenges, we propose an online optimization algorithm.

\subsubsection{Online Optimization Algorithm}
By fixing the decision variables $\{\mathbf{U}^{r}, \mathbf{p}^{r}\}_{r\in\mathcal{R}}$ and rearranging constraints (\ref{c1}) and (\ref{c9}), we have $g_{0}^{r}(\boldsymbol{\lambda}^{r}) = \frac{1}{R}\sum\nolimits_{r=1}^{R}\mathbb{E}_{\boldsymbol{\beta}^{r}}[T^{r}(\boldsymbol{\lambda}^{r})] - \gamma \leq 0$ and $g_{n}^{r}(\boldsymbol{\lambda}^{r}) = \frac{1}{R}\sum\nolimits_{r=1}^{R}\mathbb{E}_{\beta_{n}^{r}}[E_{n}^{r}(\boldsymbol{\lambda}^{r})] - \delta \leq 0$, $n\in\mathcal{N}$.
Given $\{\mathbf{U}^{r}, \mathbf{p}^{r}\}_{r\in\mathcal{R}}$, problem (\ref{opt_problem_v1}) can be transformed into the following form:
\begin{subequations}
\label{opt_problem_lambda}
\begin{alignat}{5}
&\displaystyle{\minimize_{\substack{\{\boldsymbol{\lambda}^{r}\}_{r\in\mathcal{R}}}}} & & \quad \frac{1}{R}\sum\nolimits_{r=1}^{R}g_{\mathrm{obj}}^{r}(\boldsymbol{\lambda}^{r}) \\
&\mathrm{subject\: to}
& & \quad \frac{1}{R}\sum\nolimits_{r=1}^{R}g_{0}^{r}(\boldsymbol{\lambda}^{r}) \leq 0, \label{c31}\\
& & & \quad \frac{1}{R}\sum\nolimits_{r=1}^{R}g_{n}^{r}(\boldsymbol{\lambda}^{r}) \leq 0, \:\:\: n\leq \mathcal{N}, \label{c32}\\
& & & \quad \mathrm{constraints}\:\: \textrm{(\ref{c8})},\: \textrm{(\ref{c7})}, \: \mathrm{and} \:\textrm{(\ref{c2})}.  \nonumber
\end{alignat}
\end{subequations}
By using Lyapunov optimization \cite{Neelybook}, we introduce $N+1$ virtual queues $Q_{n}^{r}$, $0\leq n \leq N$ in the $r$-th training round to account for constraints (\ref{c31}) and (\ref{c32}).
Let $\mathbf{Q}^{r} = (Q_{0}^{r}, Q_{1}^{r}, \ldots, Q_{N}^{r})\in \mathbb{R}^{N+1}$ denote the virtual queue vector.
We initialize $\mathbf{Q}^{r}$ as $\mathbf{Q}^{0} = \mathbf{0}_{N+1}$.
In the $r$-th training round, for $0 \leq n \leq N$, we update each virtual queue as
\begin{align}
    \label{eq:update_rule}
    Q_{n}^{r+1} = \max\{\mu Q_{n}^{r} + (1-\mu)g_{n}^{r}(\boldsymbol{\lambda}^{r}), 0\},
\end{align}
where $\mu \in [0, 1]$ is a tunable parameter that controls the impact of constraints (\ref{c31}) and (\ref{c32}).
We characterize the virtual queue backlog as
$L(\mathbf{Q}^{r}) = \frac{1}{2}\Vert\mathbf{Q}^{r}\Vert^{2}$, $r\in\mathcal{R}$.
We then define a Lyapunov drift to characterize the stability of the virtual queue as
\begin{align}
    \Delta^{r} = L(\mathbf{Q}^{r+1}) - L(\mathbf{Q}^{r}) = \frac{1}{2}(\Vert\mathbf{Q}^{r+1}\Vert^{2} - \Vert\mathbf{Q}^{r}\Vert^{2}), \quad r\in\mathcal{R}.
\end{align}
To jointly optimize the constraints and the objective, we define a drift-plus-penalty term as $\Delta^{r} + Vg_{\mathrm{obj}}^{r}(\boldsymbol{\lambda}^{r})$, where $V$ is a nonnegative coefficient.
To jointly guarantee the stability of the virtual queue vector and optimize our objective, instead of solving problem (\ref{opt_problem_lambda}), we solve the following optimization problem in each training round $r\in\mathcal{R}$:
\begin{subequations}
\label{opt_problem_drift}
\begin{alignat}{5}
&\displaystyle{\minimize_{\substack{\boldsymbol{\lambda}^{r}}}} & & \quad \Delta^{r} + Vg_{\mathrm{obj}}^{r}(\boldsymbol{\lambda}^{r}) \\
&\mathrm{subject\: to}
& & \quad \lambda_{1}^{r} = 1, \label{cv1} \\
& & & \quad \lambda_{m}^{r} \leq \lambda_{m-1}^{r}, \:\: m \in \mathcal{M}\backslash\{1\}, \label{cv2} \\
& & & \quad \lambda_{m}^{r} \in \{0, 1\}, \:\:  m \in \mathcal{M}\backslash\{1\}. \label{cv3}
\end{alignat}
\end{subequations}
Due to constraints \textrm{(\ref{cv1})}, (\ref{cv2}), and (\ref{cv3}), there are $M$ feasible solutions.
Hence, we can use an exhaustive search approach to solve this problem efficiently.

\subsubsection{Stability Analysis}
We provide stability analysis of our proposed online optimization algorithm and characterize the impact of the tunable parameter $\mu$.
We introduce an assumption which is widely used in the literature (e.g., \cite{9687317, he2024online}).

\noindent\textbf{Assumption 4.} 
\textit{There exist two positive constants $T^{\max}$ and $E^{\max}$ such that for arbitrary decision vector $\boldsymbol{\lambda}^{r}$, we have}
$\mathbb{E}_{\boldsymbol{\beta}^{r}}[T^{r}(\boldsymbol{\lambda}^{r})] \leq T^{\max}$ and $\mathbb{E}_{\beta_{n}^{r}}[E_{n}^{r}(\boldsymbol{\lambda}^{r})] \leq E^{\max}$, $n\in\mathcal{N}$, $r\in \mathcal{R}$.

We present three lemmas which bound the constraints, value of the virtual queue, and drift-plus-penalty term, respectively.

\noindent\textbf{Lemma 3.} 
\textit{The values of the square of constraints satisfy $(g_{0}^{r}(\boldsymbol{\lambda}^{r}))^{2} = \max\{\gamma^{2}, (T^{\max}-\gamma)^{2}\} = G_{1}$ and $(g_{n}^{r}(\boldsymbol{\lambda}^{r}))^{2} = \max\{\delta^{2}, (E^{\max}-\delta)^{2}\} = G_{2}$, $n\in\mathcal{N}$, $r\in\mathcal{R}$.}
\begin{proof}
Based on Assumption 4, we have 
\begin{align}
    \lvert g_{0}^{r}(\boldsymbol{\lambda}^{r})\rvert = \max\{\gamma, T^{\max} - \gamma\}.
\end{align}
Thus, the following equality always hold:
\begin{align}
    \big(g_{0}^{r}(\boldsymbol{\lambda}^{r})\big)^{2} = \max\{\gamma^{2}, (T^{\max}-\gamma)^{2}\}.
\end{align}
We use similar steps to obtain $\big(g_{n}^{r}(\boldsymbol{\lambda}^{r})\big)^{2} = \max\{\delta^{2}, (E^{\max}-\delta)^{2}\}$, $n\in\mathcal{N}$, $r\in\mathcal{R}$.
\end{proof}

\noindent\textbf{Lemma 4.} 
\textit{The values of virtual queues are bounded by}
$Q_{0}^{r} \leq \sqrt{G_{1}}$ and $Q_{n}^{r} \leq \sqrt{G_{2}}$, $n\in\mathcal{N}$, $r\in\mathcal{R}$.

\begin{proof}
We use induction for our proof. 
First, we have $Q_{1}^{1} = 0 < \sqrt{G_{1}}$.
Then, we suppose $Q_{0}^{r} \leq \sqrt{G_{1}}$.
$Q_{1}^{r+1}$ satisfies 
\begin{align}
    Q_{1}^{r+1} 
    \overset{(\text{a})}{\leq}&\: \lvert\mu Q_{0}^{r} + (1-\mu)g_{0}^{r}(\boldsymbol{\lambda}^{r})\rvert \nonumber\\
    \overset{(\text{b})}{\leq}&\: \mu Q_{0}^{r} + (1-\mu)\lvert g_{0}^{r}(\boldsymbol{\lambda}^{r})\rvert \nonumber\\
    \overset{(\text{c})}{\leq}&\: \mu\sqrt{G_{1}} + (1-\mu)\sqrt{G_{1}} \nonumber\\
    =&\: \sqrt{G_{1}},
\end{align}
where inequalities ($\text{a}$) and ($\text{b}$) hold due to eqn. (\ref{eq:update_rule}) and the triangle inequality, respectively.   
Inequality ($\text{c}$) results from Assumption 4 and Lemma 3.
We use similar steps to derive $Q_{n}^{r} \leq \sqrt{G_{2}}$, $n\in\mathcal{N}$, $r\in\mathcal{R}$.
\end{proof}

\noindent\textbf{Lemma 5.} 
\textit{The drift-plus-penalty term is bounded by }
\begin{align}
    &\Delta^{r} + Vg_{\mathrm{obj}}^{r}(\boldsymbol{\lambda}^{r}) \nonumber\\
    \leq&\: \frac{1}{2}(1-\mu)^{2}(G_{1} + NG_{2}) + \mu(1-\mu)(Q_{0}^{r}g_{0}^{r}(\boldsymbol{\lambda}^{r}) \nonumber\\
    &\: + \sum_{n=1}^{N}Q_{n}^{r}g_{n}^{r}(\boldsymbol{\lambda}^{r})) + Vg_{\mathrm{obj}}^{r}(\boldsymbol{\lambda}^{r}), \quad r \in \mathcal{R}.
\end{align}

\begin{proof}
The Lyapunov drift satisfies 
\begin{align}
    \Delta^{r} = \frac{1}{2}\big((Q_{1}^{r+1})^{2} - (Q_{0}^{r})^{2}\big) + \frac{1}{2}\sum_{n=1}^{N}\big((Q_{n}^{r+1})^{2} - (Q_{n}^{r})^{2}\big).
\end{align}
In particular, 
\begin{align}
    &(Q_{1}^{r+1})^{2} - (Q_{0}^{r})^{2} \nonumber\\
    \overset{(\text{a})}{=}&\: \frac{1}{2}(\mu^{2}-1)(Q_{0}^{r})^{2} + \mu(1-\mu)Q_{0}^{r}g_{0}^{r}(\boldsymbol{\lambda}^{r}) \nonumber\\
    &+ \frac{1}{2}(1-\mu)^{2}(g_{0}^{r}(\boldsymbol{\lambda}^{r}))^{2} \nonumber\\
    \overset{(\text{b})}{\leq}&\: \mu(1-\mu)Q_{0}^{r}g_{0}^{r}(\boldsymbol{\lambda}^{r}) + \frac{1}{2}(1-\mu)^{2}G_{1},
\end{align}
where equality (a) results from eqn. (\ref{eq:update_rule}).
Inequality (b) results from Lemma 3 and the fact that $\frac{1}{2}(\mu^{2}-1)(Q_{0}^{r})^{2} \leq 0$.
Then, we use similar steps to bound $(Q_{n}^{r+1})^{2} - (Q_{n}^{r})^{2}$, $n\in\mathcal{N}$.
Finally, we substitute these two terms into the drift-plus-penalty term and obtain the result.
\end{proof}

Based on Lemma 5, we present an additional assumption which is widely used in the literature (e.g. \cite{Neelybook, 9687317}).

\noindent\textbf{Assumption 5.} 
\textit{For any $r\in\mathcal{R}$, let $g_{\mathrm{obj}}^{\star}$ denote the optimal objective achieved by the decision vector $\boldsymbol{\lambda}^{r}$.
The optimal value of $g_{0}^{r}$ is denoted by $g_{0}^{\star}$.
For $n\in\mathcal{N}$, the optimal value of $g_{n}^{r}$ is denoted by $g_{n}^{\star}$.
There exists a nonnegative constant $C$ such that for an arbitrary decision vector $\boldsymbol{\lambda}^{r}$, we have}
\begin{align}
    &\Delta^{r} + Vg_{\mathrm{obj}}^{r}(\boldsymbol{\lambda}^{r}) \nonumber\\
    \leq&\: \frac{1}{2}(1-\mu)^{2}(G_{1} + NG_{2}) + \mu(1-\mu)(Q_{0}^{r}g_{0}^{\star} + \sum_{n=1}^{N}Q_{n}^{r}g_{n}^{\star}) \nonumber\\
    &+ Vg_{\mathrm{obj}}^{\star} + C.
\end{align}


Then, we show the constraint violation and performance gap in the following theorems.

\noindent\textbf{Theorem 2.} 
\textit{Based on Lemma 4, the constraint violation is bounded by}
\begin{align}
    \!\!\!\frac{1}{R}\sum_{r=1}^{R}g_{n}^{r}(\boldsymbol{\lambda}^{r}) \leq 
    \begin{cases}
        \big(1+\frac{1}{(1-\mu)R}\big)\sqrt{G_{1}}, &  \text{if $n=0$},  \\
        \big(1+\frac{1}{(1-\mu)R}\big)\sqrt{G_{2}}, &  \text{if $n\in\mathcal{N}$}.  
    \end{cases}
\end{align}

\begin{proof}
We first analyze $g_{0}^{r}(\boldsymbol{\lambda}^{r})$.
Based on eqn. (\ref{eq:update_rule}), we have 
\begin{align}
    (1-\mu)g_{0}^{r}(\boldsymbol{\lambda}^{r}) \leq Q_{1}^{r+1} - Q_{0}^{r} + (1 - \mu)Q_{0}^{r}.
\end{align}
By performing telescoping sum on both sides and using the fact that $Q_{1}^{1} = 0$, we have 
\begin{align}
    (1-\mu)\sum_{r=1}^{R}g_{0}^{r}(\boldsymbol{\lambda}^{r}) \leq Q_{1}^{R+1} + (1-\mu)\sum_{r=1}^{R}Q_{0}^{r}.
\end{align}
By dividing $(1-\mu)R$ on both sides and using Lemma 4, we have 
\begin{align}
    \frac{1}{R}\sum_{r=1}^{R}g_{0}^{r}(\boldsymbol{\lambda}^{r}) \leq \big(1+\frac{1}{(1-\mu)R}\big)\sqrt{G_{1}}.
\end{align}
We can use similar steps to obtain the result when $n\in\mathcal{N}$.
\end{proof}

\noindent\textbf{Theorem 3.} 
\textit{Based on Assumption 5 and Lemmas 3\:$-$\:4, the performance gap is bounded by}
\begin{align}
    \frac{1}{R}\sum\nolimits_{r=1}^{R}\big(g_{\mathrm{obj}}^{r}(\boldsymbol{\lambda}^{r})-g_{\mathrm{obj}}^{\star}\big) \leq \frac{W + C}{V}, \label{ineq:performance_gap}
\end{align}
\textit{where $W = \frac{1}{2}(1-\mu)^{2}(G_{1}+NG_{2}) + \mu(1-\mu)\big(\sqrt{G_{1}}(T^{\max}-\gamma) + \sqrt{G_{2}}(E^{\max}-\delta)N\big)$.}

\begin{proof}
Based on Assumption 5, we have 
\begin{align}
    &L(\mathbf{Q}^{r+1}) - L(\mathbf{Q}^{r}) + Vg_{\mathrm{obj}}^{r}(\boldsymbol{\lambda}^{r}) \nonumber\\
    \leq&\: \frac{1}{2}(1-\mu)^{2}(G_{1} + NG_{2}) + \mu(1-\mu)(Q_{0}^{r}g_{1}^{\star} + \sum_{n=1}^{N}Q_{n}^{r}g_{n}^{\star}) \nonumber\\
    &+ Vg_{\mathrm{obj}}^{\star} + C \nonumber\\
    \overset{(\text{a})}{\leq}&\: W + Vg_{\mathrm{obj}}^{\star} + C,
\end{align}
where inequality (a) results from Lemmas 3\:$-$\:4.
Then, we perform telescoping sum on both sides and obtain
\begin{align}
    &L(\mathbf{Q}^{R+1}) - L(\mathbf{Q}^{1}) + V\sum_{r=1}^{R}g_{\mathrm{obj}}^{r}(\boldsymbol{\lambda}^{r}) \nonumber\\
    \leq&\: R(W + C) + V\sum_{r=1}^{R}g_{\mathrm{obj}}^{\star}.
\end{align}

Since $L(\mathbf{Q}^{R+1}) \geq 0$ and $L(\mathbf{Q}^{1}) = 0$, we rearrange the inequality and divide both sides by $RV$ to obtain the result.
\end{proof}

\noindent\textbf{Remark 2.}
Theorems 2 and 3 suggest that as $R$ approaches infinity, the constraint violation is bounded by constant values, which indicates the stability of our proposed algorithm.
The performance gap is also bounded by a constant value. 
In addition, there is a trade-off between the performance gap and constraint violation.
That is, as $\mu$ increases, the performance gap decreases (i.e., the convergence rate is improved) while the constraint violation increases (i.e., the total delay and energy consumption on training increases) and vice versa.
We will show in Section \ref{sec:experiments} that setting $\mu$ properly can balance the learning performance and efficiency of our proposed ASFL.

\subsection{RB Allocation Subproblem}
Note that the RB allocation decision matrix $\mathbf{U}^{r}$ is independent between training rounds.
Hence, given $\mathbf{p}^{r}$ and $\boldsymbol{\lambda}^{r}$, we can optimize $\mathbf{U}^{r}$ in each training round $r\in\mathcal{R}$ independently by solving the following optimization problem:
\begin{subequations}
\label{opt_problem_RB}
\begin{alignat}{5}
&\displaystyle{\minimize_{\substack{\mathbf{U}^{r}}}} & & \quad g_{\mathrm{obj}}^{r}(\mathbf{U}^{r}) \\
&\mathrm{subject\: to}
& & \quad \mathbb{E}_{\boldsymbol{\beta}^{r}}[T^{r}(\mathbf{U}^{r})] \leq \gamma, \label{c41}\\
& & & \quad \mathbb{E}_{\beta_{n}^{r}}[E_{n}^{r}(\mathbf{U}^{r})] \leq \delta, \quad n \in \mathcal{N}, \label{c42}\\
& & & \quad \sum\nolimits_{n=1}^{N}u_{n,k}^{r} \leq 1, \quad k \in \mathcal{K}, \label{c43} \\
& & & \quad u_{n,k}^{r}\in \{0, 1\}, \quad n \in \mathcal{N}, k \in \mathcal{K}. \label{c44}
\end{alignat}
\end{subequations}
Constraints (\ref{c41}) and (\ref{c42}) are convex with respect to (w.r.t.) $\mathbf{U}^{r}$.
Problem (\ref{opt_problem_RB}) is an integer programming problem, which can be efficiently solved via a convex optimization tool (e.g., CVXPY \cite{diamond2016cvxpy}). 

\subsection{Transmit Power Allocation Subproblem}
The transmit power decision vector $\mathbf{p}^{r}$ is independent between training rounds.
Hence, given $\mathbf{U}^{r}$ and $\boldsymbol{\lambda}^{r}$, we can optimize $\mathbf{p}^{r}$ in each training round $r\in\mathcal{R}$ independently.
Tackling the delay and energy consumption constraints is challenging since they are nonconvex w.r.t. $\mathbf{p}^{r}$.
Thus, to linearize these constraints, we first introduce auxiliary variables and formulate an equivalent problem.
We then derive the solution.

We introduce three non-negative auxiliary variables (i.e., $\gamma_{1}^{r}$, $\gamma_{2}^{r}$, and $\gamma_{3}^{r}$) to bound the delay in those three stages.
Similarly, we introduce three non-negative auxiliary variables (i.e., $\delta_{1}^{r}$, $\delta_{2}^{r}$, and $\delta_{3}^{r}$) to bound $E_{n}^{\mathrm{MS}, r}$, $E_{n}^{\mathrm{UP}, r}$, and $E_{n}^{\mathrm{UP}, r}$, respectively.  
Thus, given $\mathbf{U}^{r}$ and $\boldsymbol{\lambda}^{r}$, the transmit power allocation subproblem is as follows:
\begin{subequations}
\label{opt_problem_power_v1}
\begin{alignat}{5}
&\displaystyle{\minimize_{\substack{\mathbf{p}^{r}, \gamma_{1}^{r}, \gamma_{2}^{r}, \gamma_{3}^{r},  \\ \delta_{1}^{r}, \delta_{2}^{r}, \delta_{3}^{r}}}} & & \quad g_{\mathrm{obj}}^{r}(\mathbf{p}^{r}) \\
&\mathrm{subject\: to}
& & \quad T_{n}^{\mathrm{S1},r}(p_{n}^{r}) \leq \gamma_{1}^{r}, \quad n \in \mathcal{N}, \label{c61}\\
& & & \quad T_{n}^{\mathrm{S2},r}(p_{n}^{r}) \leq \gamma_{2}^{r}, \quad n \in \mathcal{N}, \label{c62}\\
& & & \quad \mathbb{E}_{\beta_{n}^{r}}[T_{n}^{\mathrm{S3},r}(p_{n}^{r})] \leq \gamma_{3}^{r}, \quad n \in \mathcal{N}, \label{c63}\\
& & & \quad E_{n}^{\mathrm{MS}, r}(p_{n}^{r}) \leq \delta_{1}^{r}, \quad n \in \mathcal{N}, \label{c68}\\
& & & \quad E_{n}^{\mathrm{UP}, r}(p_{n}^{r}) \leq \delta_{2}^{r}, \quad n \in \mathcal{N}, \label{c64}\\
& & & \quad \mathbb{E}_{\beta_{n}^{r}}[E_{n}^{\mathrm{CBP}, r}(p_{n}^{r})] \leq \delta_{3}^{r}, \quad n \in \mathcal{N}, \label{c65}\\
& & & \quad \gamma_{1}^{r} + \gamma_{2}^{r} + \gamma_{3}^{r} \leq \gamma, \label{c66}\\
& & & \quad \delta_{1}^{r} + \delta_{2}^{r} + \delta_{3}^{r} \leq \delta - \max\nolimits_{n\in\mathcal{N}}\{E_{n}^{\mathrm{FP}, r}\},  \label{c67}\\
& & & \quad 0 \leq p_{n}^{r} \leq P^{\mathrm{max}}, \quad n \in \mathcal{N}. \label{c69}
\end{alignat}
\end{subequations}
To solve problem (\ref{opt_problem_power_v1}), we introduce an iterative algorithm to optimize $\mathbf{p}^{r}$ and $\{\gamma_{1}^{r}, \gamma_{2}^{r}, \gamma_{3}^{r}, \delta_{1}^{r}, \delta_{2}^{r}, \delta_{3}^{r}\}$ alternately.
First, given $\{\gamma_{1}^{r}, \gamma_{2}^{r}, \gamma_{3}^{r}, \delta_{1}^{r}, \delta_{2}^{r}, \delta_{3}^{r}\}$, 
we determine the feasible domain w.r.t. $\mathbf{p}^{r}$ based on constraints (\ref{c61})\:$-$\:(\ref{c65}).
From constraint (\ref{c61}), $p_{n}^{r}$ satisfies
\begin{align}
\label{ineq:p_c1}
    p_{n}^{r} \geq \frac{BN_{0}\bigg(2^{{\frac{(\boldsymbol{\lambda}^{r-1} - \boldsymbol{\lambda}^{r})^{\intercal}\boldsymbol{\psi}}{\gamma_{1}^{r}B\sum_{k=1}^{K}u_{n,k}^{r}}}}-1\bigg)}{{\lvert h_{n}^{r}\rvert^{2}}}.
\end{align}
From constraint (\ref{c62}), $p_{n}^{r}$ satisfies
\begin{align}
\label{ineq:p_c2}
    \!p_{n}^{r} \geq \frac{BN_{0}\bigg(2^{{\frac{f_{n}\sum_{m=1}^{M-1}(\lambda_{m}^{r} - \lambda_{m+1}^{r})q_{m}}{(\gamma_{2}^{r}f_{n}-\kappa^{\mathrm{c}}D_{n}B(\boldsymbol{\lambda}^{r})^{\intercal}\mathbf{v}^{\mathrm{FP}})\sum_{k=1}^{K}u_{n,k}^{r}}}}\!-\!1\bigg)}{\lvert h_{n}^{r}\rvert^{2}}.
\end{align}
From constraint (\ref{c63}), $p_{n}^{r}$ satisfies
\begin{align}
\label{ineq:p_c3}
    p_{n}^{r} \leq \frac{1}{\ln{\Bigg(\frac{\omega_{1,n}^{r} - \gamma_{3}^{r}}{\omega_{1,i}^{r}\mathbb{E}_{\lvert h_{n}^{r}\rvert}\big[\exp\big(-\frac{\alpha BN_{0}\sum_{k=1}^{K}u_{n,k}^{r}}{\lvert h_{n}^{r}\rvert^{2}}\big)\big]}\Bigg)}},
\end{align}
where 
\begin{align}
    \omega_{1,n}^{r} =&\: \frac{\kappa^{\mathrm{s}}D_{n}(\boldsymbol{\mathbf{1}}_{M}-\boldsymbol{\lambda}^{r})^{\intercal}\big(\mathbf{v}^{\mathrm{FP}}+\mathbf{v}^{\mathrm{BP}}\big)}{f^{\mathrm{s}}} + \frac{\kappa^{\mathrm{c}}D_{n}(\boldsymbol{\lambda}^{r})^{\intercal}\mathbf{v}^{\mathrm{BP}}}{f_{n}} \nonumber\\
    &+ \frac{\sum_{m=1}^{M-1}(\lambda_{m}^{r} - \lambda_{m+1}^{r})\psi_{m+1}}{B^{\mathrm{DN}}\log_{2}(1 + \frac{p^{\mathrm{s}}\lvert h_{n}^{r}\rvert^{2}}{B^{\mathrm{DN}}N_{0}})}.
\end{align}
For constraint (\ref{c68}), it is equivalent to $\frac{\ln\big(1 + \frac{p_{n}^{r}\lvert h_{n}^{r}\rvert^{2}}{BN_{0}}\big)}{p_{n}^{r}} \geq \frac{(\boldsymbol{\lambda}^{r-1} - \boldsymbol{\lambda}^{r})^{\intercal}\boldsymbol{\psi}\ln{2}}{\delta_{1}^{r}B\sum_{k=1}^{K}u_{n,k}^{r}}$.
We use the Taylor series to expand the numerator on the left-hand side of this inequality. 
In particular, $p_{n}^{r}$ satisfies
\begin{align}
\label{ineq:p_c8}
    p_{n}^{r} \leq \frac{2BN_{0}}{\lvert h_{n}^{r}\rvert^{2}}\left(1 - \frac{(\boldsymbol{\lambda}^{r-1} - \boldsymbol{\lambda}^{r})^{\intercal}\boldsymbol{\psi}N_{0}\ln{2}}{\delta_{1}^{r}\sum_{k=1}^{K}u_{n,k}^{r}\lvert h_{n}^{r}\rvert^{2}}\right).
\end{align}
Similarly, from constraint (\ref{c64}), $p_{n}^{r}$ satisfies
\begin{align}
\label{ineq:p_c4}
    \!\!\!p_{n}^{r} \leq \frac{2BN_{0}}{\lvert h_{n}^{r}\rvert^{2}}\left(1\!-\! \frac{D_{n}N_{0}\ln{2}\sum_{m=1}^{M-1}(\lambda_{m}\!- \!\lambda_{m+1})q_{m}}{\delta_{2}^{r}\sum_{k=1}^{K}u_{n,k}^{r}\lvert h_{n}^{r}\rvert^{2}}\right).
\end{align}
From constraint (\ref{c65}), $p_{n}^{r}$ satisfies
\begin{align}
\label{ineq:p_c5}
    p_{n}^{r} 
    \leq
    \frac{1}{\ln\left(\frac{D_{n}(\boldsymbol{\lambda}^{r})^{\intercal}\mathbf{v}^{\mathrm{BP}}\phi\kappa^{\mathrm{c}}(f_{n})^{2}\sum_{k}u_{n,k}^{r}\:-\:\delta_{3}^{r}}{\mathbb{E}_{\lvert h_{n}^{r}\rvert}\big[\mathrm{exp}\big(-\frac{\alpha BN_{0}\sum_{k}u_{n,k}^{r}}{\lvert h_{n}^{r}\rvert^{2}}\big)\big]D_{n}(\boldsymbol{\lambda}^{r})^{\intercal}\mathbf{v}^{\mathrm{BP}}\phi\kappa^{\mathrm{c}}(f_{n})^{2}}\right)}.
\end{align}

Let $\{p_{n}^{\mathrm{c1},r}, p_{n}^{\mathrm{c2},r}, p_{n}^{\mathrm{c3},r}, p_{n}^{\mathrm{c4},r}, p_{n}^{\mathrm{c5},r}, p_{n}^{\mathrm{c6},r}\}_{n\in\mathcal{N}}$ denote the right-hand side of inequalities (\ref{ineq:p_c1})\:$-$\:(\ref{ineq:p_c5}), respectively.
Given $\{\gamma_{1}^{r}, \gamma_{2}^{r}, \gamma_{3}^{r}, \delta_{1}^{r}, \delta_{2}^{r}, \delta_{3}^{r}\}$, problem (\ref{opt_problem_power_v1}) can be reformulated as
\begin{subequations}
\label{opt_problem_power_v2}
\begin{alignat}{5}
&\displaystyle{\minimize_{\substack{\mathbf{p}^{r}}}} & & \quad g_{\mathrm{obj}}^{r}(\mathbf{p}^{r}) \\
&\mathrm{subject\: to}
& & \quad \max\{p_{n}^{\mathrm{c1},r}, p_{n}^{\mathrm{c2},r}\} \leq p_{n}^{r} \leq  \min\{P^{\mathrm{max}}, p_{n}^{\mathrm{c3},r},\nonumber\\
& & & \quad p_{n}^{\mathrm{c4},r}, p_{n}^{\mathrm{c5},r}, p_{n}^{\mathrm{c6},r}\}, \quad n\in\mathcal{N}.
\end{alignat}
\end{subequations}
The solution to problem (\ref{opt_problem_power_v2}) is given by Theorem 4.

\noindent\textbf{Theorem 4. }{Given $\{\gamma_{1}^{r}, \gamma_{2}^{r}, \gamma_{3}^{r}, \delta_{1}^{r}, \delta_{2}^{r}, \delta_{3}^{r}\}$, for $n\in\mathcal{N}$, the solution to problem (\ref{opt_problem_power_v2}) is obtained as}
\begin{equation}
\label{eq:opt_power}
    \!\!\!p_{n}^{r} = 
    \begin{cases}
        \max\{p_{n}^{\mathrm{c1},r}, p_{n}^{\mathrm{c2},r}\}, & \!\text{C1},  \\
        \frac{1}{\ln{(-\frac{\omega_{3,n}^{r}}{2\omega_{2,n}^{r}})}}, & \!\text{C2},\\
        \min\{P^{\mathrm{max}}, p_{n}^{\mathrm{c3},r}, p_{n}^{\mathrm{c4},r}, p_{n}^{\mathrm{c5},r}, p_{n}^{\mathrm{c6},r}\}, & 
        \!\text{otherwise},
    \end{cases} 
\end{equation}
where $\omega_{2, n}^{r} = \mathbb{E}_{\lvert h_{n}^{r}\rvert}\big[\exp \big(-\frac{\alpha BN_{0}\sum_{k=1}^{K}u_{n,k}^{r}}{\lvert h_{n}^{r}\rvert^{2}}\big)\big]^{2}\Vert\mathbf{w}_{n, \boldsymbol{\lambda}^{r}}^{\mathrm{s}, r}\Vert^{2}$, 
$\omega_{3, n}^{r} = -2\mathbb{E}_{\lvert h_{n}^{r}\rvert}\big[\exp \big(-\frac{\alpha BN_{0}\sum_{k=1}^{K}u_{n,k}^{r}}{\lvert h_{n}^{r}\rvert^{2}}\big)\big]\langle\bar{\mathbf{w}}_{\boldsymbol{\lambda}^{r}}^{\mathrm{s}, r},  \mathbf{w}_{n, \boldsymbol{\lambda}^{r}}^{\mathrm{s}, r}\rangle$, $n\in\mathcal{N}$,
C1 is the condition where $\omega_{3, n}^{r} < 0$ and $1/\ln{(-\frac{\omega_{3,n}^{r}}{2\omega_{2,n}^{r}})} < \max\{p_{n}^{\mathrm{c1},r}, p_{n}^{\mathrm{c2},r}\}$. 
C2 is the condition where $\omega_{3, n}^{r} < 0$ and $\max\{p_{n}^{\mathrm{c1},r}, p_{n}^{\mathrm{c2},r}\} \leq 1/\ln{(-\frac{\omega_{3,n}^{r}}{2\omega_{2,n}^{r}})} \leq \min\{P^{\mathrm{max}}, p_{n}^{\mathrm{c3},r}, p_{n}^{\mathrm{c4},r}, p_{n}^{\mathrm{c5},r}, p_{n}^{\mathrm{c6},r}\}$.
    
\begin{proof}
We first determine the critical point of $g_{\mathrm{obj}}^{r}(\mathbf{p}^{r})$ as $1/\ln{(-\frac{\omega_{3,n}^{r}}{2\omega_{2,n}^{r}})}$.
Under C1, $g_{\mathrm{obj}}^{r}(\mathbf{p}^{r})$ is increasing and is minimized at the lower bound of the domain.
Under C2, $g_{\mathrm{obj}}^{r}(\mathbf{p}^{r})$ is minimized at the critical point.
Otherwise, $g_{\mathrm{obj}}^{r}(\mathbf{p}^{r})$ is decreasing and is minimized at the upper bound of the domain.
\end{proof}

\begin{algorithm}[t] 
    \small
    \caption{Iterative Algorithm for Problem (\ref{opt_problem_power_v2})}
    \label{alg:power}   
    \begin{algorithmic}[1] 
        \STATE \textbf{Input}: Iteration index $\tau_{1} := 0$, power objective tolerance $\varepsilon_{\mathrm{p}}$.

        \STATE Randomly initialize $\{\gamma_{1}^{r, 0}, \gamma_{2}^{r, 0}, \gamma_{3}^{r, 0}, \delta_{1}^{r, 0}, \delta_{2}^{r, 0}, \delta_{3}^{r, 0}\}$.
  
        \STATE \textbf{While} $\tau_{1} = 0\:\mathrm{or}\: \lvert g_{\mathrm{p}}(\mathbf{p}^{r,  \tau_{1}}) - g_{\mathrm{p}}(\mathbf{p}^{r,  \tau_{1}-1})\rvert > \varepsilon_{\mathrm{p}}$\: \textbf{do}
        \begin{ALC@g}
            \STATE $\tau_{1} := \tau_{1} + 1$.
            
            \STATE 
            Update $\mathbf{p}^{r,  \tau_{1}}$ based on eqn. (\ref{eq:opt_power}).

            \STATE 
            Update $\{\gamma_{1}^{r, \tau_{1}}, \gamma_{2}^{r, \tau_{1}}, \gamma_{3}^{r, \tau_{1}}, 
            \delta_{1}^{r, \tau_{1}}, 
            \delta_{2}^{r, \tau_{1}}, \delta_{3}^{r, \tau_{1}}\}$.


        \end{ALC@g}
        \STATE \textbf{End while}
  
        \STATE \textbf{Output}: $\mathbf{p}^{r, \tau_{1}}$.
  
    \end{algorithmic}\normalsize
    \end{algorithm}

Second, given $\mathbf{p}^{r}$, we update $\{\gamma_{1}^{r}, \gamma_{2}^{r}, \gamma_{3}^{r}, \delta_{1}^{r}, \delta_{2}^{r}, \delta_{3}^{r}\}$ 
as
$\gamma_{1}^{r} = \max_{n\in\mathcal{N}}\{T_{n}^{\mathrm{S1},r}(p_{n}^{r})\}$, 
$\gamma_{2}^{r} = \max_{n\in\mathcal{N}}\{T_{n}^{\mathrm{S2},r}(p_{n}^{r})\}$,
$\gamma_{3}^{r} = \max_{n\in\mathcal{N}}\{\mathbb{E}_{\beta_{n}^{r}}[T_{n}^{\mathrm{S3},r}(p_{n}^{r})]\}$,
$\delta_{1}^{r} = \max_{n\in\mathcal{N}}\{E_{n}^{\mathrm{MS}, r}(p_{n}^{r})\}$, 
$\delta_{2}^{r} = \max_{n\in\mathcal{N}}\{E_{n}^{\mathrm{UP}, r}(p_{n}^{r})\}$, 
and $\delta_{3}^{r} = \max_{n\in\mathcal{N}}\{\mathbb{E}_{\beta_{n}^{r}}[E_{n}^{\mathrm{CBP}, r}(p_{n}^{r})]\}$, respectively.
This iterative algorithm is presented in Algorithm \ref{alg:power}.

In summary, to solve problem (\ref{opt_problem_v1}), 
we update each decision variable by solving problems (\ref{opt_problem_drift}), (\ref{opt_problem_RB}), and (\ref{opt_problem_power_v2}) alternately in each training round.
Our proposed OOE-BCD algorithm is shown in Algorithm \ref{alg:BCD}.

\subsection{Computation Complexity Analysis}
To determine the computation complexity of our proposed OOE-BCD algorithm in each training round, we analyze each of the three subproblems.
For the model splitting subproblem, since there are at most $M$ feasible solutions, the computation complexity is linear in the number of model splitting decisions, which is $\mathcal{O}(M)$.
For the RB allocation subproblem, the computation complexity is $\mathcal{O}(N^{K})$.
For the transmit power allocation subproblem, since we need to determine the transmit power for each client, the computation complexity is $\mathcal{O}(N\log_{2}(\frac{1}{\varepsilon_{\mathrm{p}}}))$.
Therefore, the overall computation complexity of our proposed OOE-BCD algorithm in one training round is $\mathcal{O}(\log_{2}(\frac{1}{\varepsilon_{\mathrm{o}}})(M + N^{K} + N\log_{2}(\frac{1}{\varepsilon_{\mathrm{p}}}))$.

\begin{algorithm}[t] 
    \small
    \caption{OOE-BCD Algorithm for Problem (\ref{opt_problem_v1})}
    \label{alg:BCD}   
    \begin{algorithmic}[1] 
        \STATE \textbf{Input}: Iteration index $\tau_{2} := 0$, objective tolerance $\varepsilon_{\mathrm{o}}$.

        \STATE Randomly initialize $\{ \mathbf{U}^{r, 0}, \mathbf{p}^{r, 0}\}$.
  
        \STATE \textbf{While} $\tau_{2} = 0\:\mathrm{or}\: \lvert g_{\mathrm{obj}}^{r}(\mathbf{U}^{r, \tau_{2}}, \mathbf{p}^{r, \tau_{2}}, \boldsymbol{\lambda}^{r, \tau_{2}}) - g_{\mathrm{obj}}^{r}(\mathbf{U}^{r, \tau_{2}-1}, \mathbf{p}^{r, \tau_{2}-1}, \boldsymbol{\lambda}^{r, \tau_{2}-1})\rvert > \varepsilon_{\mathrm{o}}$ \textbf{do}
        \begin{ALC@g}
            \STATE $\tau_{2} := \tau_{2} + 1$.
        
            \STATE 
            Update $\boldsymbol{\lambda}^{r, \tau_{2}}$ by solving problem (\ref{opt_problem_drift}).

            \STATE 
            Update $\mathbf{U}^{r, \tau_{2}}$ by solving problem (\ref{opt_problem_RB}).

            \STATE 
            Update $\mathbf{p}^{r, \tau_{2}}: = \mathbf{p}^{r, \tau_{1}}$ based on Alg. \ref{alg:power}.


        \end{ALC@g}
        \STATE \textbf{End while}
  
        \STATE \textbf{Output}: $\{\mathbf{U}^{r, \star}, \mathbf{p}^{r, \star}, \boldsymbol{\lambda}^{r, \star}\} := \{\mathbf{U}^{r, \tau_{2}}, \mathbf{p}^{r, \tau_{2}}, \boldsymbol{\lambda}^{r, \tau_{2}}\}$.
  
    \end{algorithmic}\normalsize
\end{algorithm}


\begin{table}[t]
\centering
\caption{List of key simulation parameters}
\label{tb:para_table}
\resizebox{0.95\columnwidth}{!}{%
\begin{tabular}{|c|c|c|c|c|c|}
\hline
\textbf{Para.} & \textbf{Value} & \textbf{Para.} & \textbf{Value} & \textbf{Para.} & \textbf{Value}\\
\hline
$B$ & 1 MHz  & $P^{\max}$ & 1.5 W  & $K$ & 8 \\
\hline
$B^{\mathrm{DN}}$ & 8 MHz & $\kappa^{\mathrm{s}}$ & $\frac{1}{32}$ cycles/FLOP  & $p^{\mathrm{s}}$ & 5 W \\
\hline
$N_{0}$ & $-$173 dBm/Hz  & $\kappa^{\mathrm{c}}$ & $\frac{1}{16}$ cycles/FLOP  & $\iota$ & 0.05 \\
\hline
$\alpha$ & 1  & $f^{\mathrm{s}}$ & $10^{10}$ cycles/s & $V$ & 10 \\
\hline
$\gamma$ & 20 s  & $\delta$ & $0.5$ J & $\phi$ & $10^{-28}$ \\
\hline
\end{tabular}
}
\end{table} 

\section{Performance Evaluation}
\label{sec:performance_evaluation} 

\subsection{Simulation Setup}
We consider a scenario where clients are randomly located in a circular coverage area with a radius of 500 m.
The central server is located at the center of the area.
The channel coefficient $h_{n}^{r}$ is obtained as $h_{n}^{r} = \sqrt{\theta_{n}^{r}}\chi_{n}^{r}$,
where $\theta_{n}^{r} = -30 - 40\log_{10}(d_{n}^{r})$ denotes the path loss coefficient (in dB) \cite{10274134} and $d_{n}^{r}$ denotes the distance between client $n\in\mathcal{N}$ and the central server in the $r$-th training round.
$\chi_{n}^{r} \sim \mathcal{CN}(0,1)$ denotes the Rayleigh fading coefficient.
The CPU frequency of each client $n \in \mathcal{N}$ follows $f_{n}\sim\mathcal{U}[1\,\mathrm{GHz},\, 1.6\,\mathrm{GHz}]$ \cite{lin2023efficient}. 
We set $\varepsilon_{\mathrm{o}} = \varepsilon_{\mathrm{p}} = 0.01$.
Unless stated otherwise, we set $\mu = 0.5$.
A list of key simulation parameters is presented in Table \ref{tb:para_table}.

We use CIFAR-10 and CIFAR-100 \cite{krizhevsky2009learning} as the datasets. 
We adopt VGG-19 \cite{simonyan2014very} and ResNet-50 \cite{he2016deep} as the models with 19 and 50 layers, respectively.
The model is allowed to be split only before the convolutional layer or the fully-connected layer.
The number of training rounds $R$, the learning rate $\eta$, and the batch size are set to 200, 0.0001, and 64, respectively.
To study the effect of non-independent and identically distributed (non-IID) data partitioning, similar to other works in FL \cite{chen2020fedbe, 9835537}, we use the Dirichlet distribution $\mathrm{Dir}(\rho)$ with parameter $\rho$ to distribute the data with different labels across clients.
Let $N_{\mathrm{c}}$ denote the number of classes of a dataset.
The Dirichlet distribution $\text{Dir}(\rho)$ generates the probability vector $\mathbf{q}_n = (q^1_n, q^2_n, \ldots, q^{N_c}_n)$, where $q^{i}_{n}$ represents the fraction of data samples from class $i$ assigned to client $n$, with $\sum_{n=1}^{N_c} q^i_n = 1$. 
The probability density function of the Dirichlet distribution is given by
$f(\mathbf{q}_n; \rho) = \frac{\Gamma(N_c \rho)}{\left( \Gamma(\rho) \right)^{N_c}} \prod_{n=1}^{N_c} (q^i_n)^{\rho - 1}$,
where $\Gamma(\cdot)$ is the gamma function.
When $\rho$ decreases, the data heterogeneity across clients increases and vice versa.
Unless stated otherwise, we set $N=10$ and $\rho = 10$.
To ensure fairness, we allocate the same number of training samples to each client's local dataset.
We compare the performance of our proposed ASFL with the following five baseline schemes:
\begin{itemize}
    \item FedAvg \cite{mcmahan2017communication}: Clients train their local models simultaneously.
    We use the same RB and transmit power allocation decisions as ASFL.

    \item SL \cite{vepakomma2018split}: Clients perform training sequentially.
    We use the same RB and transmit power allocation decisions as ASFL and randomly make the model splitting decision.

    \item SFL \cite{thapa2022splitfed}: Clients train their client-side models in parallel and send the updated client-side models to the central server to perform model aggregation.
    We randomly make the model splitting decision.
    Then, we use the same RB and transmit power allocation decisions as ASFL.

    \item ACC-SFL \cite{10304624}: The model splitting and RB allocation decisions are determined using its proposed optimization algorithm. 
    We randomly choose each client's transmit power.

    \item EPSL \cite{lin2023efficient}: The model splitting and resource allocation decisions are made using its proposed optimization algorithm. 
\end{itemize}

{We introduce evaluation metrics for our experiments as follows:}
\begin{itemize}
    \item Average testing accuracy:
    This is the ratio of the total number of correct predictions to the number of testing samples.
    It quantifies how well the trained model performs on downstream tasks.

    \item Total delay on training:
    This is the end-to-end model training latency.

    \item Total energy consumption on training: 
    This metric captures the total communication and computation energy consumption of all clients.
    It corresponds to the total battery usage of end devices.

    \item Average packet error rate: 
    This is the ratio of the number of packets with errors to the total number of transmitted packets.
    It captures the communication reliability over wireless links, which strongly affects the success rate of model update.
\end{itemize}

\subsection{Experiments}
\label{sec:experiments}
\begin{table}[t]
\centering
\caption{Comparison of the average testing accuracy between different baseline schemes.}
\label{tb:acc}
\setlength{\extrarowheight}{1ex}
\resizebox{0.88\columnwidth}{!}{
\begin{tabular}{c|c|cc}
\specialrule{1.5pt}{0pt}{0pt}
\textbf{Base Model} & \textbf{Baseline}  & \textbf{CIFAR-10 (\%)} &
\textbf{CIFAR-100 (\%)} \\
\specialrule{1.5pt}{0pt}{0pt}
\multirow{6}{*}{VGG-19} 
& ASFL & 89.20 & 62.10  \\
\cline{2-4}
& FedAvg \cite{mcmahan2017communication} & 89.69 & 61.35 \\
\cline{2-4}
& SL \cite{vepakomma2018split} & 83.41 & 58.30 \\
\cline{2-4}
& SFL \cite{thapa2022splitfed} & 88.77 & 62.43 \\
\cline{2-4}
& ACC-SFL \cite{10304624} & 88.02 & 61.96 \\
\cline{2-4}
& EPSL \cite{lin2023efficient} & 87.59 & 61.58 \\
\specialrule{1.5pt}{0pt}{0pt}
\multirow{6}{*}{ResNet-50} 
& ASFL & 92.52 & 66.19  \\
\cline{2-4}
& FedAvg \cite{mcmahan2017communication} & 90.91 & 65.72 \\
\cline{2-4}
& SL \cite{vepakomma2018split} & 88.81 & 62.27 \\
\cline{2-4}
& SFL \cite{thapa2022splitfed} & 91.07 & 65.92 \\
\cline{2-4}
& ACC-SFL \cite{10304624} & 90.94 & 64.08 \\
\cline{2-4}
& EPSL \cite{lin2023efficient} & 89.47 & 63.46 \\
\specialrule{1.5pt}{0pt}{0pt}
\end{tabular}}
\end{table}

\begin{figure}[t]
    \centering
    \subfigure[]{\includegraphics[width=0.23\textwidth]{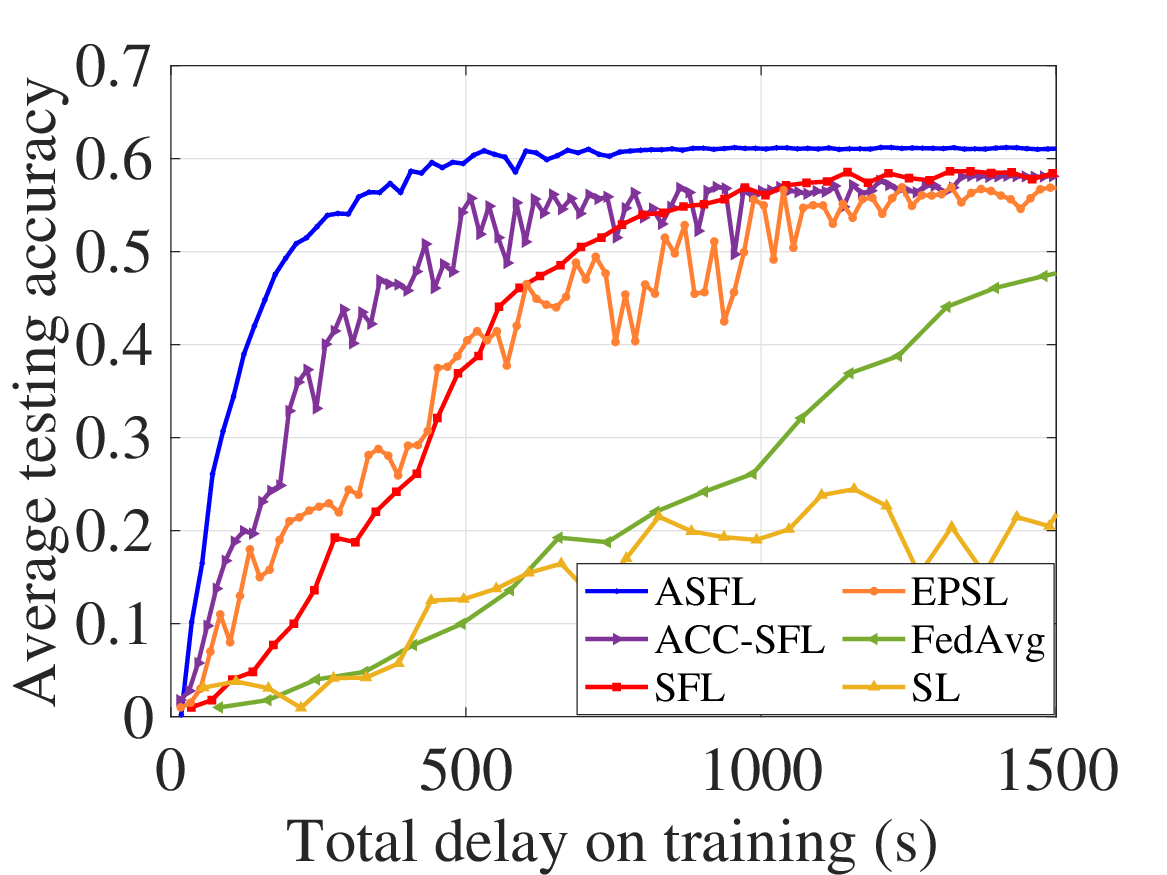}} 
    \subfigure[]{\includegraphics[width=0.23\textwidth]{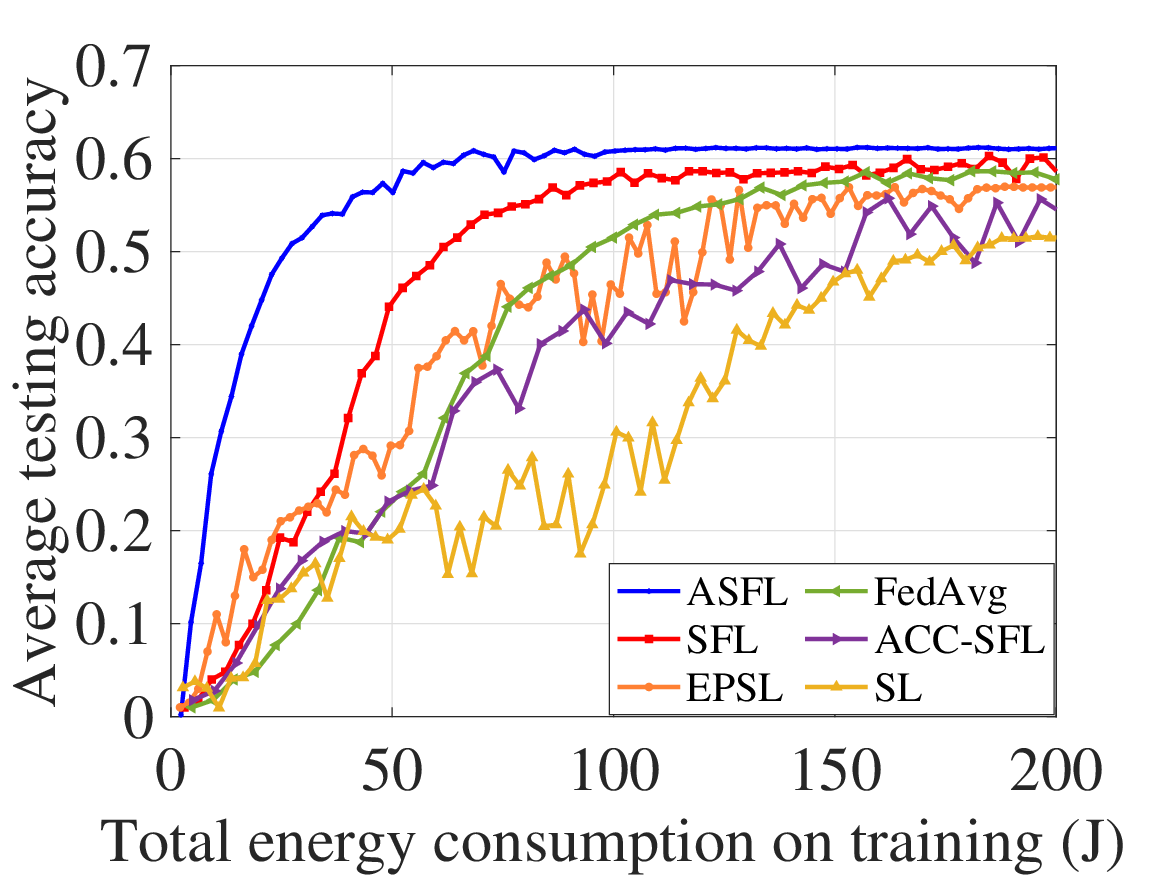}} 
    \caption{Comparison of the average testing accuracy versus (a) total delay and (b) total energy consumption on training.}
    \label{fig:acc}
\end{figure}

\begin{figure}[t]
    \centering
    \begin{minipage}[b]{0.48\columnwidth}
        \centering
        \includegraphics[width=\columnwidth]{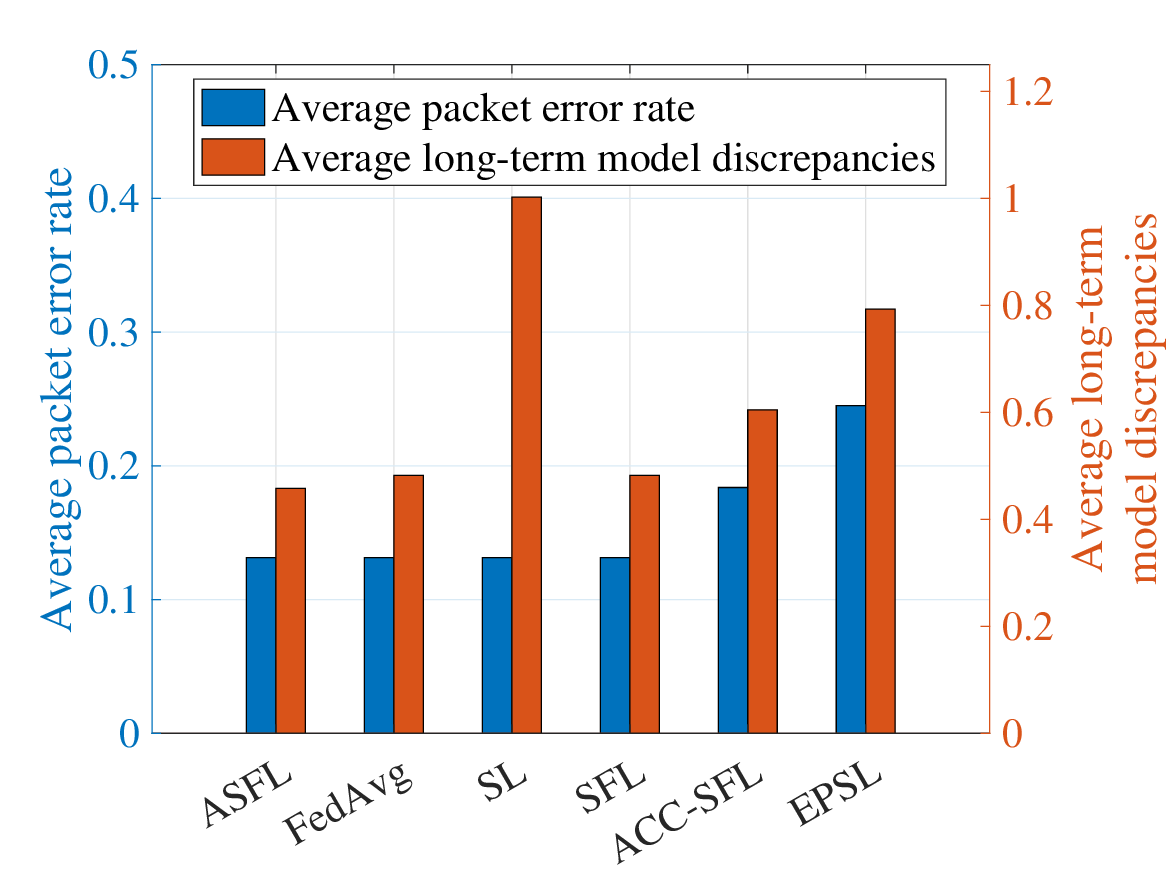} 
        \caption{Comparison of the average packet error rate and average long-term model discrepancies.}
        \label{fig:dis_err}
    \end{minipage}
    \hfill
    \begin{minipage}[b]{0.48\columnwidth}
        \centering
        \includegraphics[width=\columnwidth]{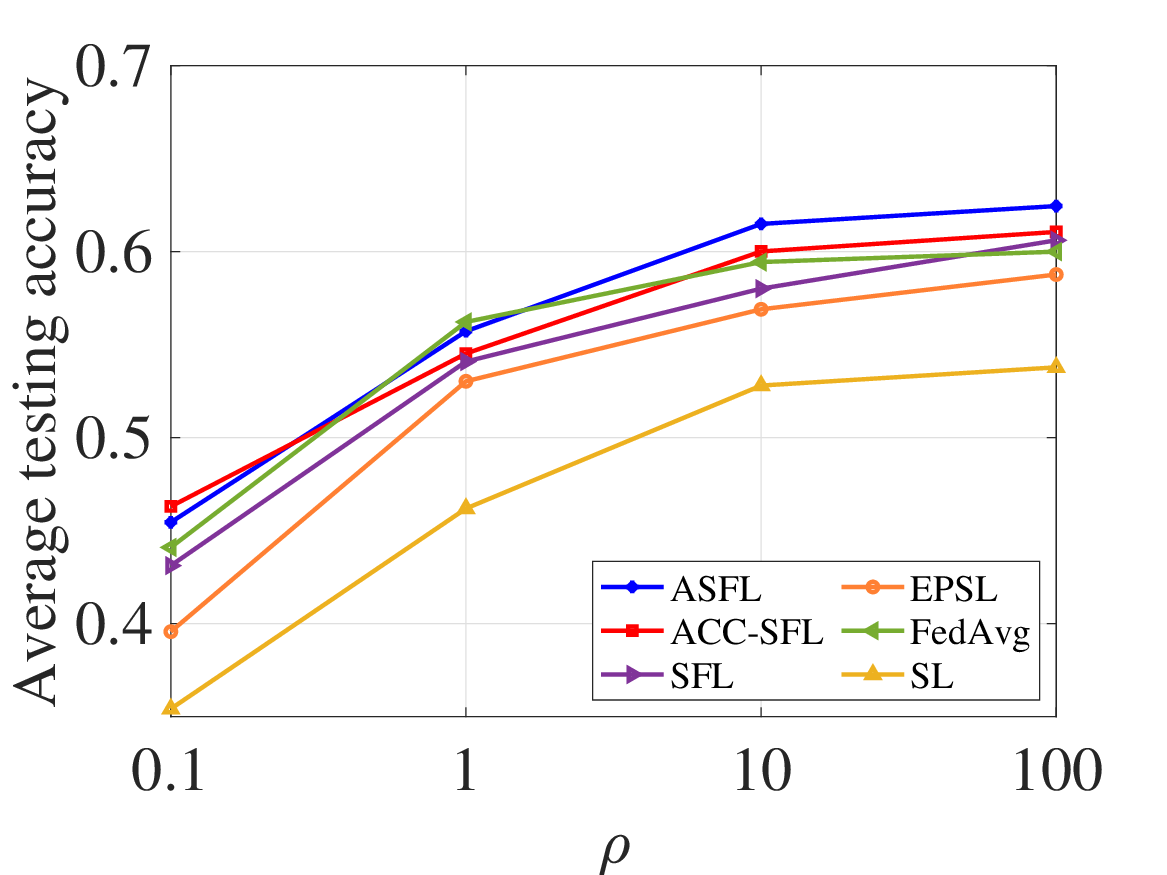} 
        \caption{Effect of the degree of data heterogeneity $\rho$ on the average testing accuracy.}
    \label{fig:rho}
    \end{minipage}
\end{figure}

\begin{figure}[t]
    \centering
    \begin{minipage}[b]{0.48\columnwidth}
        \centering
        \includegraphics[width=\columnwidth]{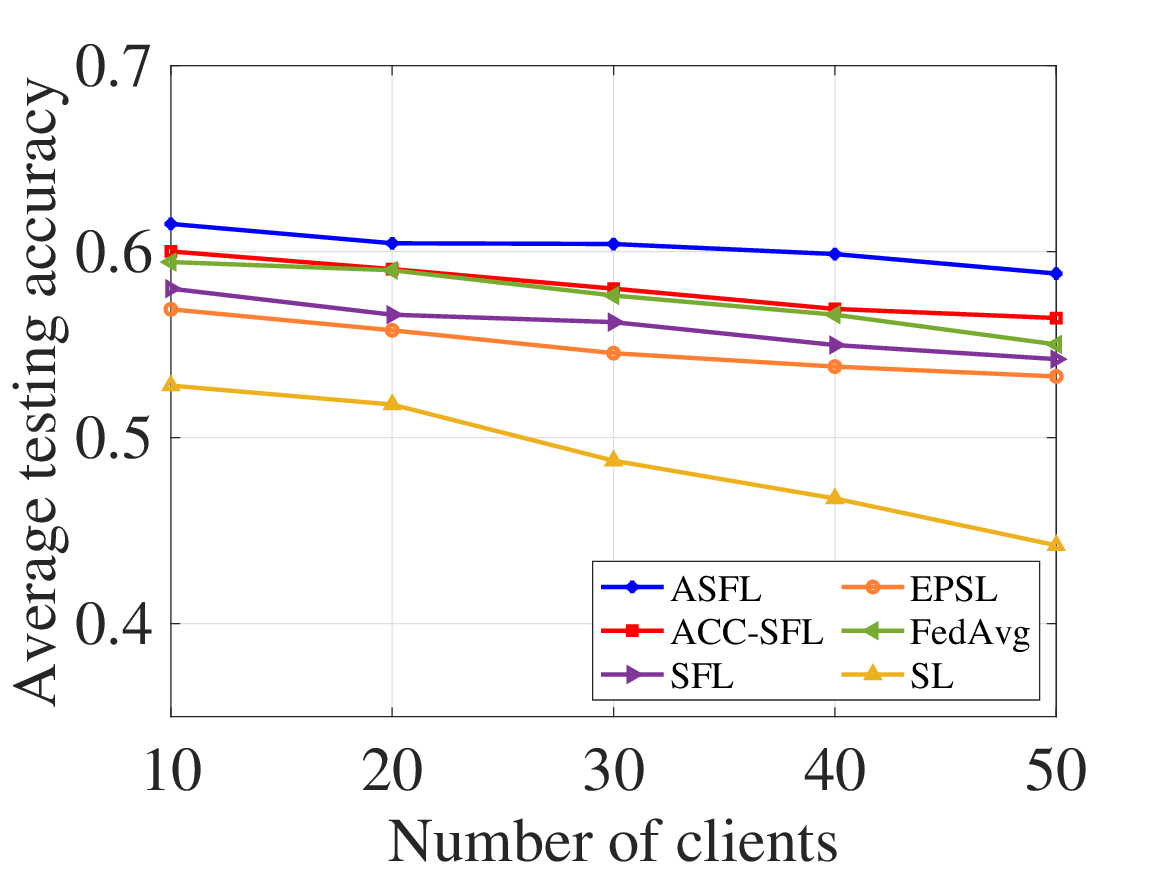} 
        \caption{Effect of the number of clients on the average testing accuracy.}
        \label{fig:num_clients}
    \end{minipage}
    \hfill
    \begin{minipage}[b]{0.48\columnwidth}
        \centering
        \includegraphics[width=\columnwidth]{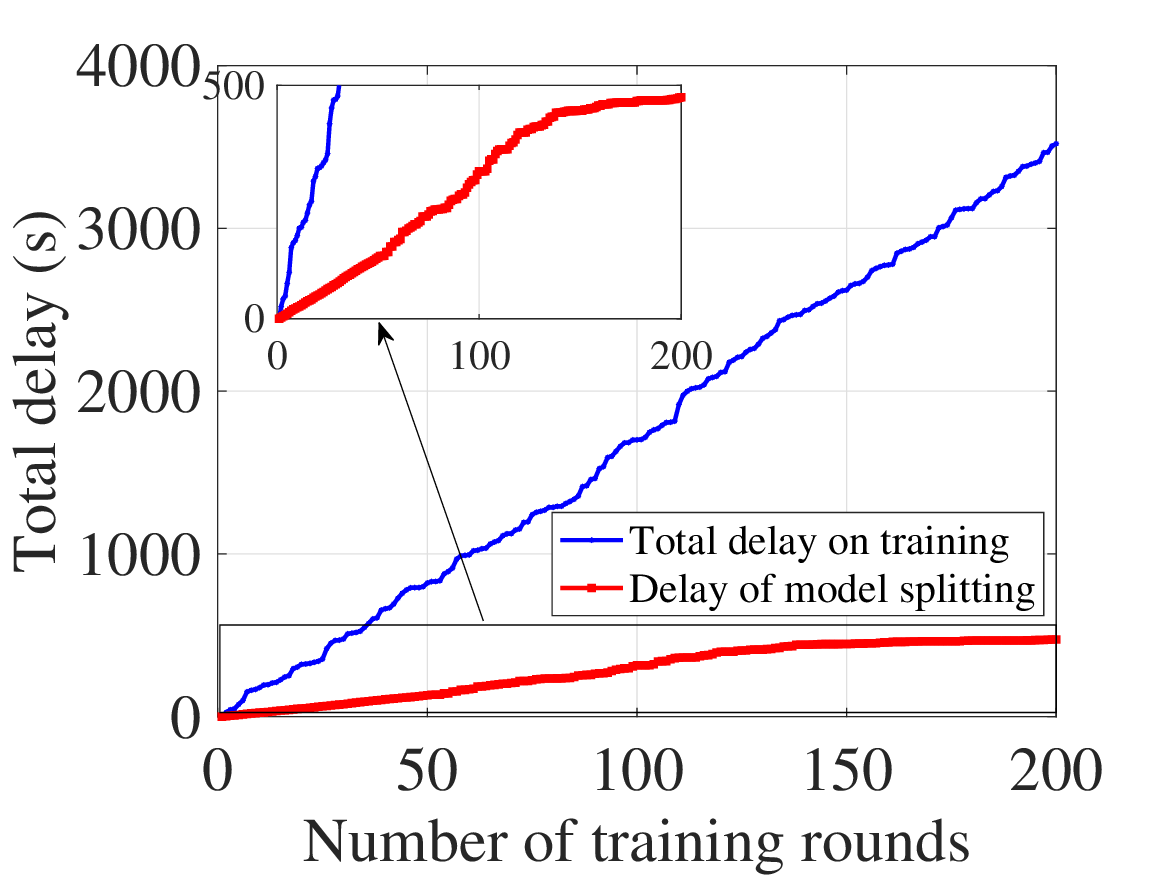} 
        \caption{Comparison of the total delay on training and model splitting.}
    \label{fig:delay_model_splitting}
    \end{minipage}
\end{figure}

\subsubsection{Comparison of the Learning Performance and Efficiency}
{In Table III, we compare the average testing accuracy on CIFAR-10 and CIFAR-100 datasets.
Results show that our proposed ASFL outperforms the baseline schemes in most cases.
In particular, when testing under ResNet-50 with CIFAR-100 dataset, our proposed ASFL outperforms FedAvg, SL, SFL, ACC-SFL, and EPSL in terms of the average testing accuracy by 1.61\%, 3.71\%, 1.45\%, 1.58\%, and 3.05\%, respectively.
This validates the effectiveness of ASFL in improving the learning performance.}

To compare the learning efficiency, we compare the average testing accuracy using VGG-19 and CIFAR-100 under the first 1500 s and 200 J in Fig. \ref{fig:acc}. 
Results show that our proposed ASFL converges faster and achieves a higher average testing accuracy and a lower total delay and energy consumption when compared with the baseline schemes.
In particular, our proposed ASFL outperforms FedAvg and SL in both the total delay and energy consumption on training.
In addition, to achieve an average testing accuracy of 0.5, our proposed ASFL achieves a total delay that is 51\%, 69\%, and 75\% lower than ACC-SFL, SFL, and EPSL, respectively.
Our proposed ASFL also reduces the energy consumption by 80\%, 56\%, and 74\% when compared with ACC-SFL, SFL, and EPSL, respectively.
It shows that our proposed ASFL improves the learning efficiency. 

\subsubsection{Comparison of the Average Packet Error Rate and Average Long-Term Model Discrepancies}
To further evaluate the learning performance, we compare the average packet error rate of all clients of our proposed ASFL with the baseline schemes using VGG-19 and CIFAR-100 in Fig. \ref{fig:dis_err}.
Results show that our proposed ASFL achieves a lower average packet error rate when compared with ACC-SFL and EPSL.
Note that our proposed ASFL achieves the same packet error rate as FedAvg, SL, and SFL since they use the same RB and transmit power allocation decisions.
In addition, our proposed algorithm achieves a lower average long-term model discrepancies when compared with the baseline schemes.
The lower average packet error rates and reduced average long-term model discrepancies contribute to the better learning performance of our proposed ASFL.

\subsubsection{Effect of Data Heterogeneity}
{We evaluate the robustness of the proposed ASFL algorithm under different levels of data heterogeneity across clients by comparing the average testing accuracy across different values of $\rho$ using VGG-19 and CIFAR-100.
A smaller $\rho$ corresponds to a higher degree of non-IID data distribution.
To control the degree of data heterogeneity across clients, we set $\rho$ to be 0.1, 1, 10, and 100.
As shown in Fig. \ref{fig:rho}, ASFL outperforms the baseline schemes in terms of the average testing accuracy in most cases.
These results highlight the robustness of ASFL to different degrees of non-IID client data distributions.}

\subsubsection{Scalability Analysis}
{We investigate the scalability of our proposed ASFL by setting the number of clients to 10, 20, 30, 40, and 50, respectively.
We compare the average testing accuracy achieved by our proposed ASFL framework with the baseline schemes using VGG-19 and CIFAR-100 in Fig. \ref{fig:num_clients}.
It can be observed that our proposed AFSL framework constantly outperforms the baseline schemes, validating the scalability of our proposed ASFL.}

\subsubsection{Overhead of Adaptive Model Splitting}
{In Fig. \ref{fig:delay_model_splitting}, we present the comparison between the total delay on training and the delay incurred by adaptive model splitting using VGG-19 and CIFAR-100. 
It can be observed that the overhead introduced by switching the model splitting points contributes only a small fraction of the overall training delay. 
In particular, at the end of the 200th training round, the total delay of adaptive model splitting accounts for 20.86\% of the total delay on training.}

{Furthermore, the model splitting points are switched almost every round in the first 150 training rounds.
This is necessary to maintain the convergence and adapt to the dynamic wireless channel conditions.
Thus, the corresponding delay of model splitting increases.
Then, as training progresses and the model begins to converge, the need for splitting point switching diminishes, resulting in a significantly reduced frequency of model splitting switching in the subsequent stages.
Therefore, the delay of model splitting gradually stabilizes.}

\begin{figure}[t]
    \centering
    \subfigure[]{\includegraphics[width=0.23\textwidth]{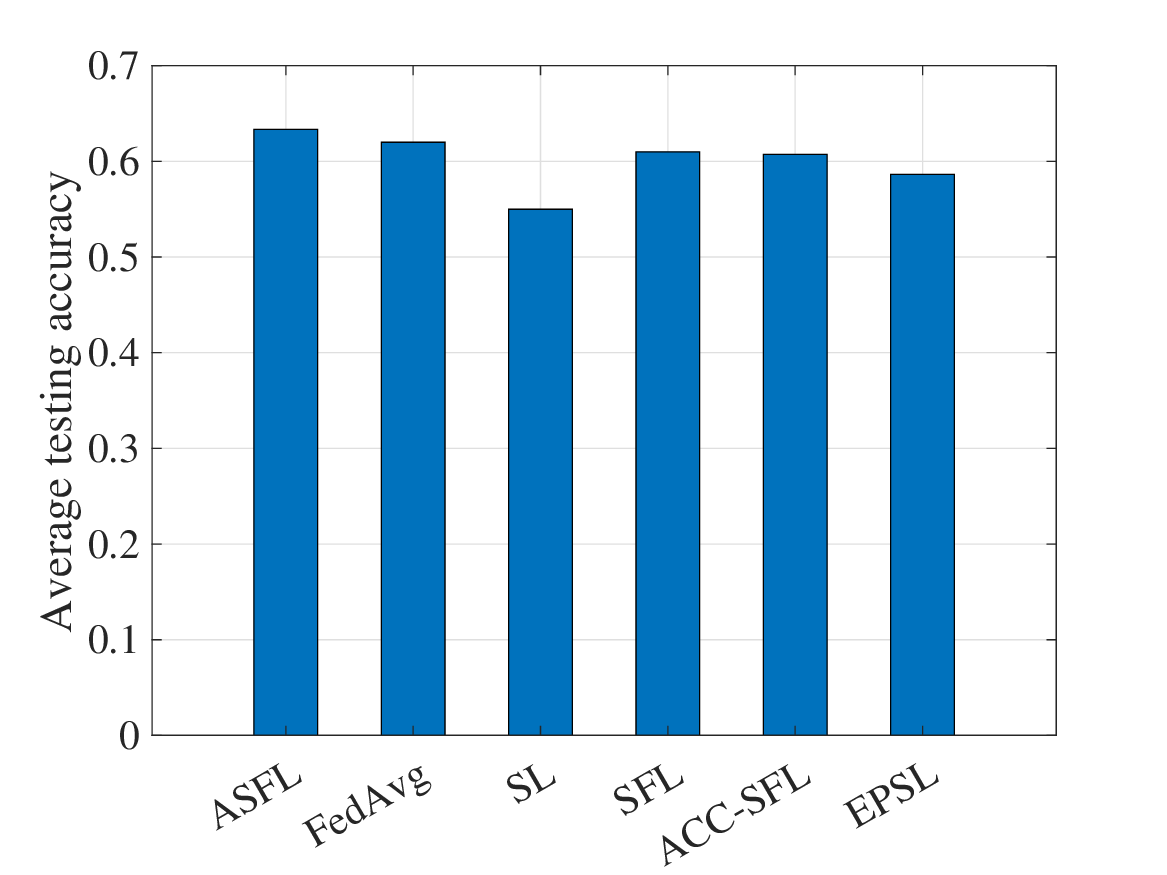}} 
    \subfigure[]{\includegraphics[width=0.23\textwidth]{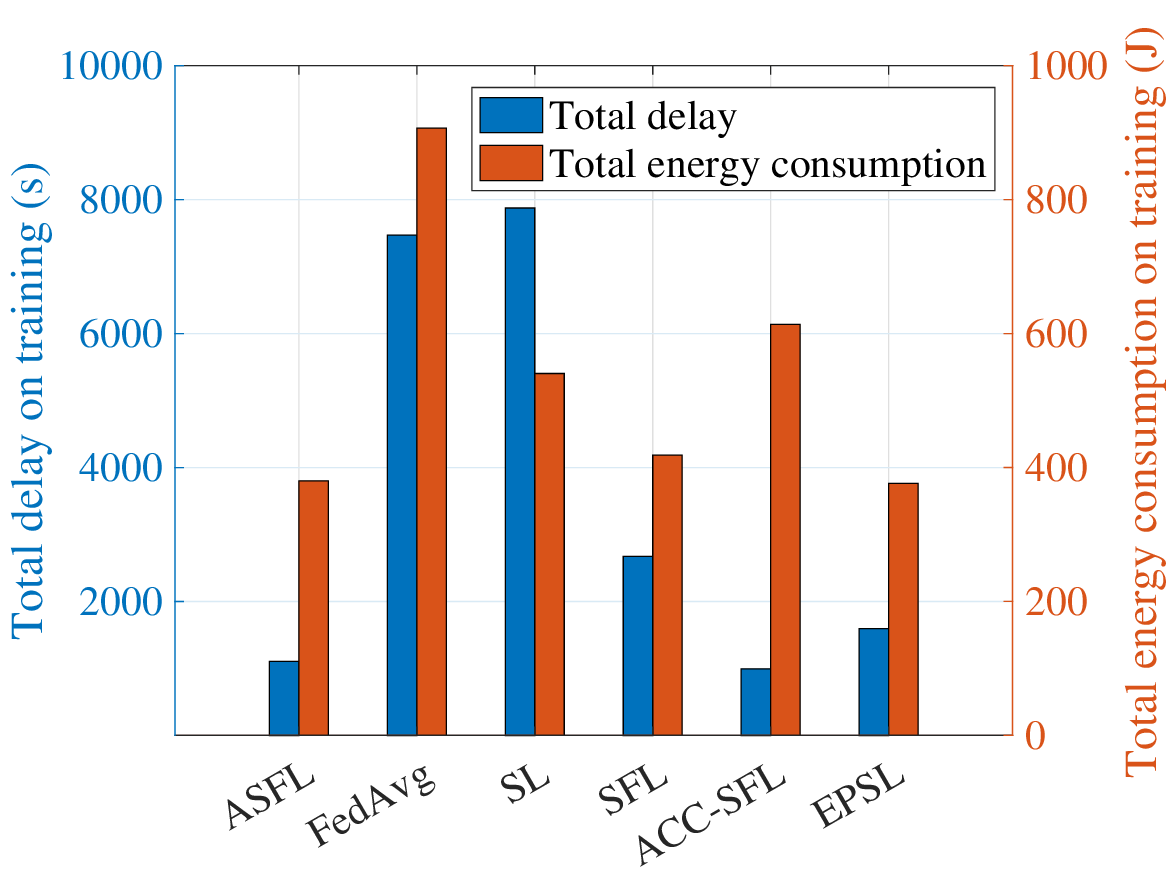}} 
    \caption{Comparison of (a) average testing accuracy and (b) total delay and energy consumption on training.}
    \label{fig:diff_scenario}
\end{figure} 

\subsubsection{Experiment on the Real-world Environment}
{To show the performance of our proposed ASFL on real-world scenarios, we use RENEW/FDD Massive MIMO dataset \cite{8368089}, which measured channel state information (CSI) collected in outdoor environments to conduct the experiments.
Since these measurements are taken over real radio links, they capture the impact of fading and interference on the channel quality.
For each client in our experiments, we calculate the channel gain by selecting a specific transmit–receive antenna pair from the measured CSI tensors in the dataset.}

{In Fig. \ref{fig:diff_scenario}, we present the comparison of the average testing accuracy, total delay on training, and total energy consumption on training using VGG-19 and CIFAR-100.
We can observe that our proposed ASFL outperforms the baseline schemes, showcasing the effectiveness of ASFL in improving the learning performance.
In addition, our proposed ASFL achieves the lowest total delay and energy consumption on training when compared with the baseline schemes, indicating the effectiveness of ASFL in improving the learning efficiency.}

\begin{figure}[t]
    \centering
    \subfigure[]{\includegraphics[width=0.23\textwidth]{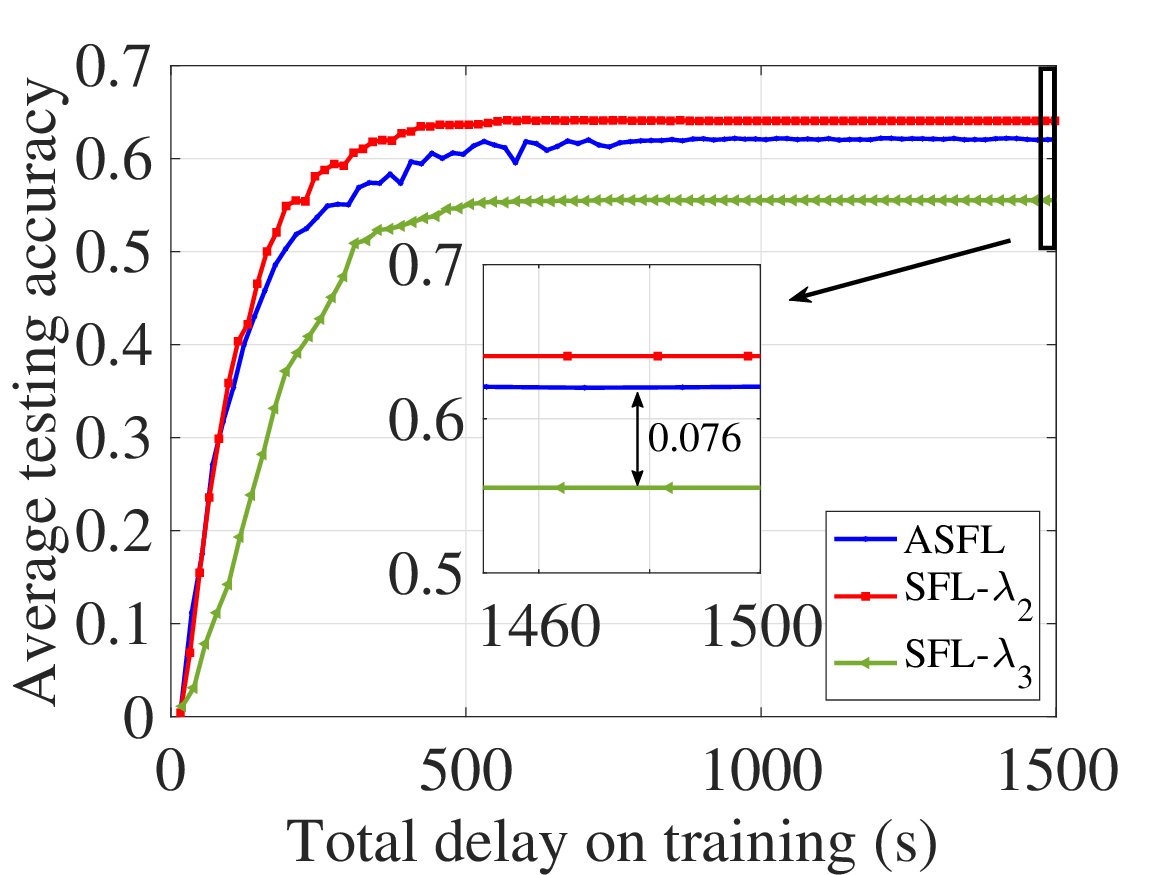}} 
    \subfigure[]{\includegraphics[width=0.23\textwidth]{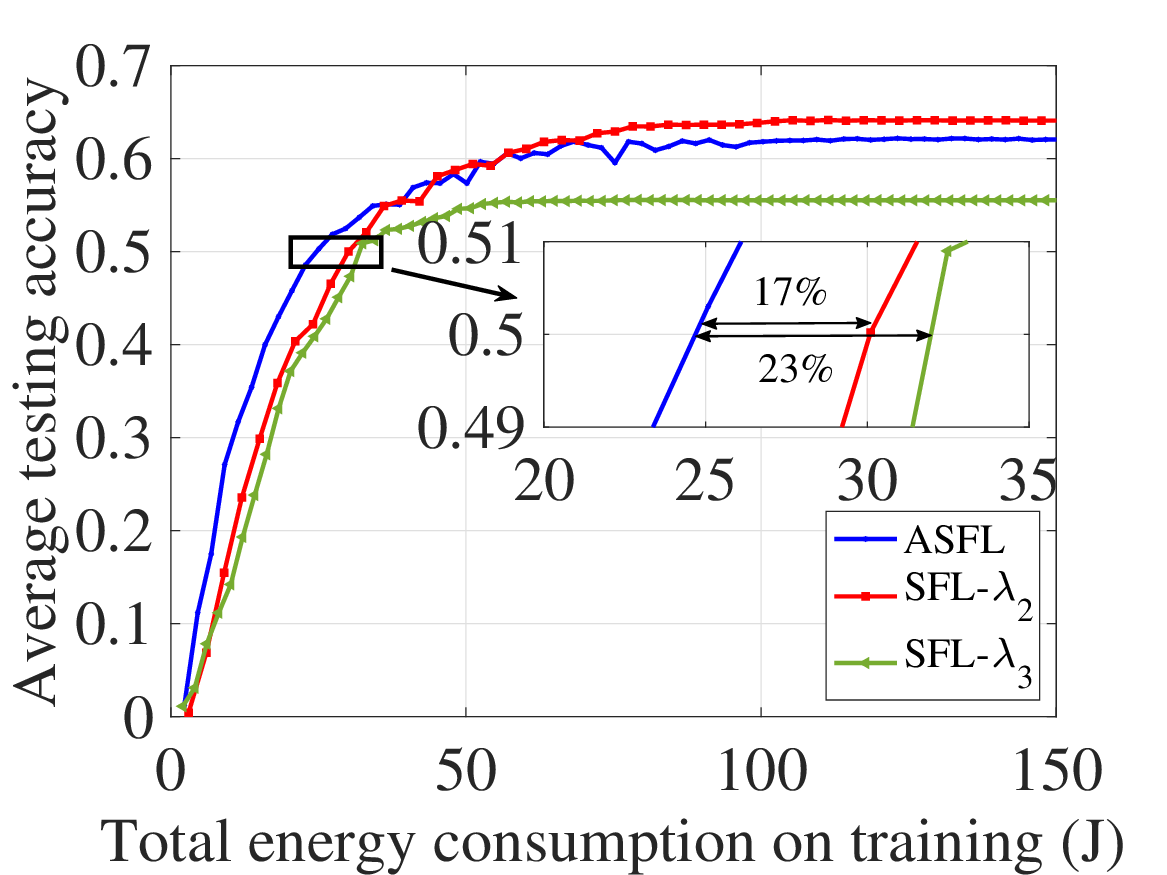}} 
    \caption{Effect of the adaptive model splitting on the average testing accuracy versus (a) total delay and (b) total energy consumption on training.}
    \label{fig:compare_MS}
\end{figure} 

\begin{figure}[t]
    \centering
    \subfigure[]{\includegraphics[width=0.23\textwidth]{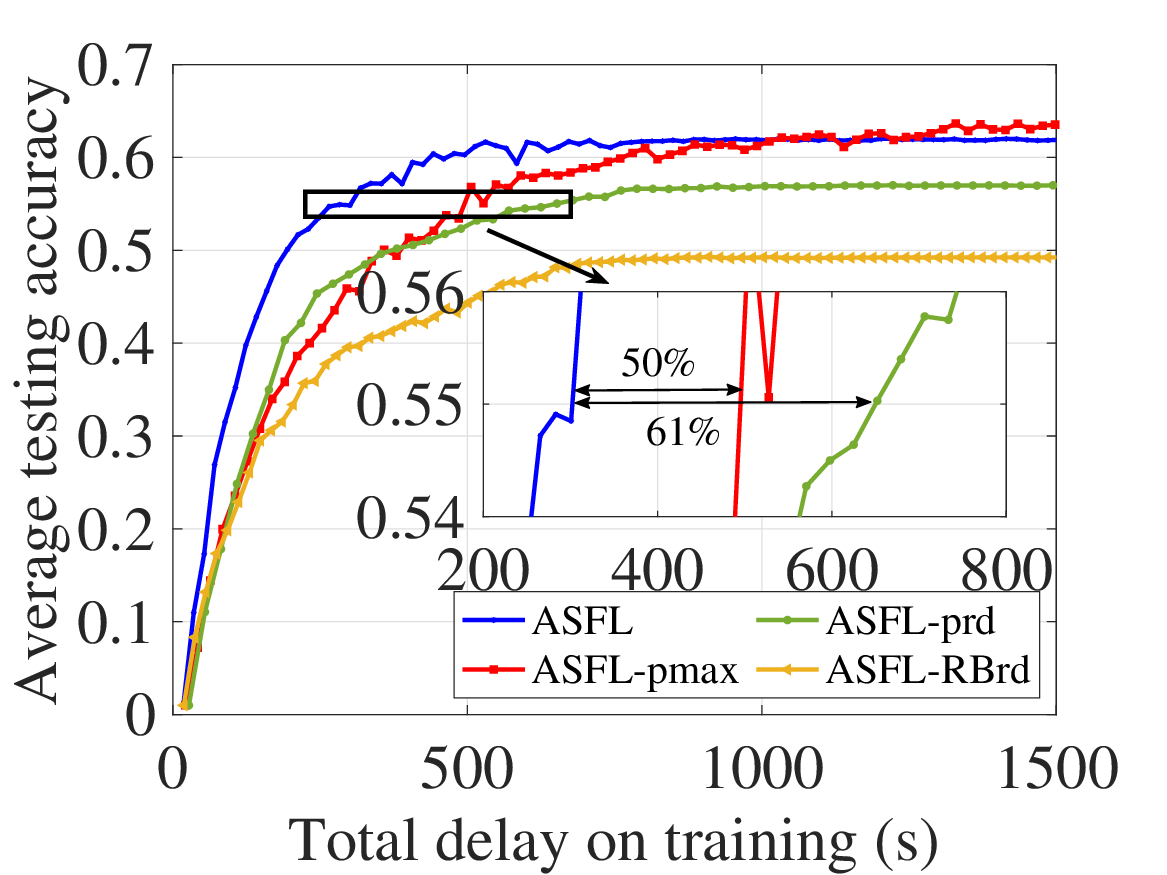}} 
    \subfigure[]{\includegraphics[width=0.23\textwidth]{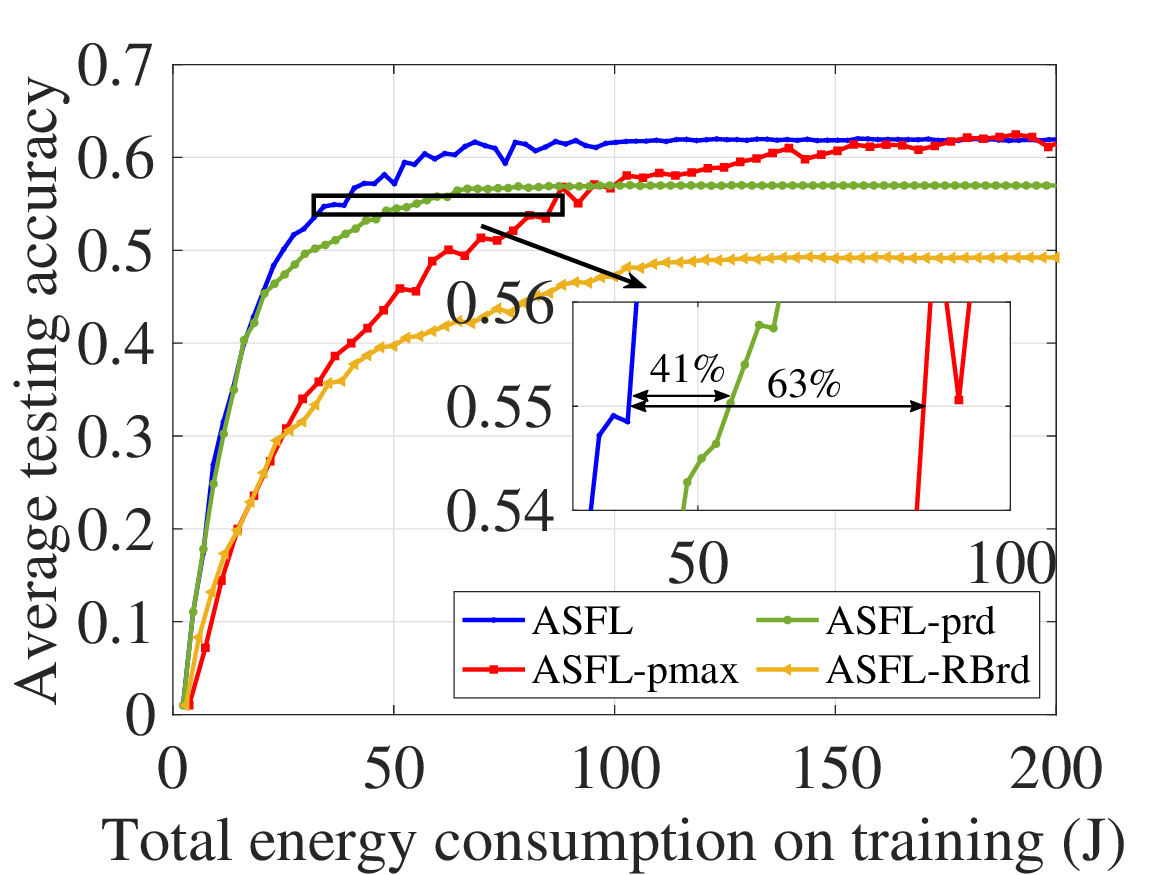}} 
    \caption{Effect of the RB and transmit power allocation on the average testing accuracy versus (a) total delay and (b) total energy consumption on training.}
    \label{fig:compare_power}
\end{figure}

\subsubsection{Effect of Adaptive Model Splitting}   
We also consider a baseline scheme (SFL-$\lambda_{m}$) by pre-splitting the model at the $m$-th layer before training and optimizing other decision variables.
We choose $m = 2$ and $3$ since they can achieve higher average testing accuracies.
Fig. \ref{fig:compare_MS} shows that using VGG-19 and CIFAR-100, our proposed ASFL converges faster than SFL-$\lambda_{3}$ and achieves an average testing accuracy that is 0.076 higher than SFL-$\lambda_{3}$.
Our proposed ASFL also reduces the total energy consumption in early training rounds.
For example, to achieve an average testing accuracy of 0.5, our proposed ASFL achieves a total energy consumption on training that is 17\% and 23\% lower than SFL-$\lambda_{2}$ and SFL-$\lambda_{3}$, respectively.
It indicates the effectiveness of adaptive model splitting.


\subsubsection{Effect of the RB Allocation and Transmit Power Allocation}
We consider three other baseline schemes: (a) ASFL-pmax: we set clients' transmit power to the maximum value and optimize other decision variables, (b) ASFL-prd: we randomly set clients' transmit power and optimize other decision variables, and (c) ASFL-RBrd: we randomly allocate RBs to clients and optimize other decision variables.
Fig. \ref{fig:compare_power} shows that using VGG-19 and CIFAR-100, our proposed ASFL outperforms ASFL-RBrd in both the total delay and energy consumption on training.
In addition, to achieve an average testing accuracy of 0.55, ASFL achieves a total delay on training that is 50\% and 61\% lower than ASFL-pmax and ASFL-prd, respectively.
ASFL also reduces the energy consumption by 63\% and 41\% when compared with ASFL-pmax and ASFL-prd, respectively.

\begin{figure}[t]
    \centering
    \subfigure[]{\includegraphics[width=0.15\textwidth]{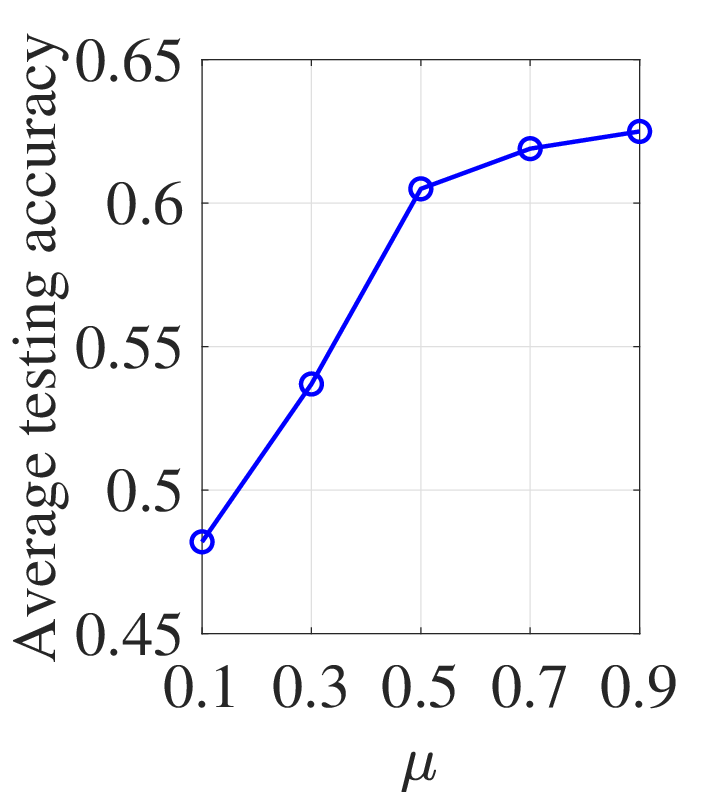}} 
    \subfigure[]{\includegraphics[width=0.15\textwidth]{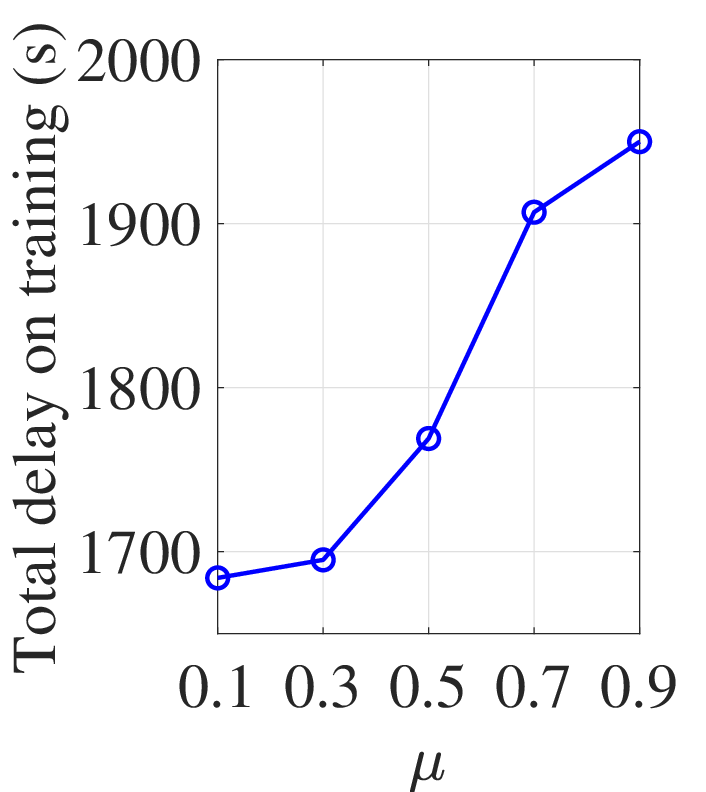}} 
    \subfigure[]{\includegraphics[width=0.15\textwidth]{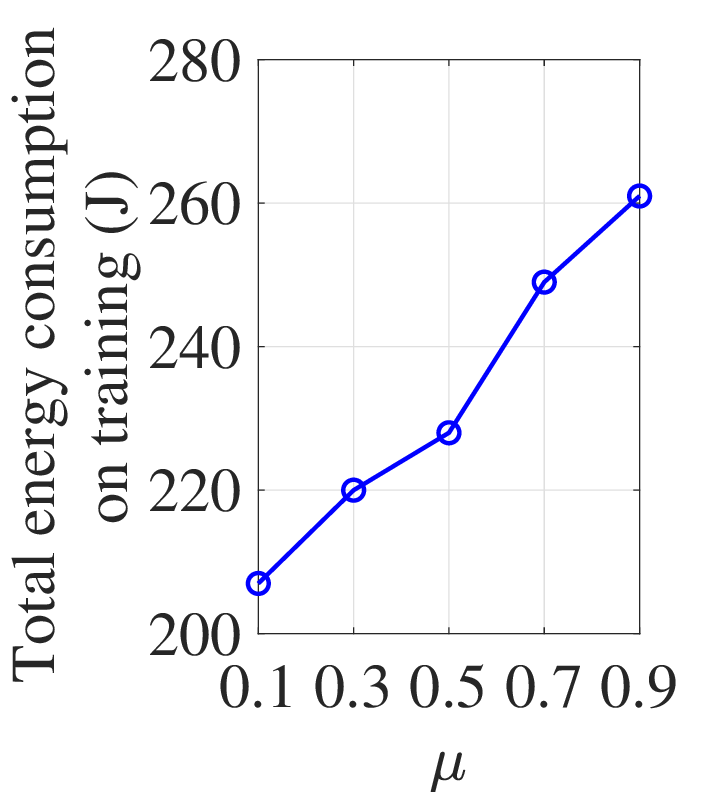}}
    \caption{Effect of $\mu$ on the (a) average testing accuracy, (b) total delay, and (c) total energy consumption on training.}
    \label{fig:mu}
\end{figure} 

\subsubsection{Effect of the Tunable Parameter $\mu$}
\label{sec:mu}
Fig. \ref{fig:mu} shows the effect of $\mu$ on the average testing accuracy, total delay and energy consumption of 100 training rounds using VGG-19 and CIFAR-100.
When $\mu$ varies from 0.1 to 0.9, the average testing accuracy increases with an increase in the total delay and energy consumption.
This is because as $\mu$ increases, the corresponding weights on constraints (\ref{c31}) and (\ref{c32}) become smaller. 
Constraints (\ref{c31}) and (\ref{c32}) have less impact and become larger when solving problem (\ref{opt_problem_drift}).
It indicates the trade-off of tuning $\mu$ on the learning performance and efficiency.

\section{Conclusion}
\label{sec:conclusion}
In this paper, we proposed an ASFL framework over wireless networks.
It enables adaptive model splitting and provides efficient resource allocation during training.
We theoretically analyzed the convergence rate of our proposed ASFL framework.
We designed an OOE-BCD algorithm to adaptively determine the model splitting and resource allocation decisions.
Experimental results showed that when compared with five baseline schemes, our proposed ASFL framework converged faster and reduced the total delay and energy consumption on training by up to 75\% and 80\%, respectively.
We also showed the robustness of our proposed algorithm to different degrees of data heterogeneity across clients' local datasets.
For future work, we plan to design ASFL with heterogeneous model splitting to further improve the learning efficiency under heterogeneous hardware configurations across clients.

\bibliography{egbib}
\end{document}